\documentclass[a4paper]{article}

\usepackage{chngcntr}   
\usepackage[numbers]{natbib}  
\usepackage{subfigure}
\usepackage[english]{babel}
\usepackage[utf8x]{inputenc}
\usepackage[T1]{fontenc}

\usepackage[a4paper,top=3cm,bottom=2cm,left=3cm,right=3cm,marginparwidth=1.75cm]{geometry}
\usepackage[colorinlistoftodos]{todonotes}
\usepackage[colorlinks=true, allcolors=blue]{hyperref}

\usepackage{enumitem}

\usepackage{authblk} 
\usepackage{setspace}
\usepackage{amsfonts}
\usepackage{bm}
\usepackage{xspace}
\usepackage{amsthm}
\usepackage{dsfont}
\usepackage{algorithm,rotating}
\usepackage{graphicx} 
\usepackage{url}
\usepackage{xcolor,soul}
\usepackage{thm-restate}
\usepackage[small,bf]{caption}
\usepackage{standalone}
\usepackage{verbatim}
\usepackage{setspace}
\usepackage{mdwmath}
\usepackage{amssymb}
\usepackage{amsmath}
\usepackage{xfrac}
\usepackage{mathtools}  
\usepackage{chngcntr}
\usepackage{tabulary}
\usepackage{booktabs}
\usepackage{epsfig}
\usepackage{wrapfig}
\usepackage{lipsum}
\usepackage{caption}
\usepackage{multirow}
\usepackage{algcompatible} 
\usepackage{footnote}
\usepackage{hyperref}

\newtheorem{theorem}{Theorem}

\newtheorem{lemma}{Lemma}

\newtheorem{cor}{Corollary}

\newcommand{\sgn}{\operatorname{sgn}}
\newcommand{\indi}{\mathds{1}}

\newcommand{\R}{\mathbb{R}}

\newcommand{\diag}{\operatorname{diag}}

\newcommand{\PP}{\mathbb{P}}
\newcommand{\vecx}{\mathbf{x}}
\newcommand{\vecw}{\mathbf{w}}

\newcommand{\vecv}{\mathbf{v}}

\newcommand{\vecu}{\mathbf{u}}

\newcommand{\vecz}{\mathbf{z}}

\newcommand{\matP}{\mathbf{P}}

\newcommand{\matW}{\mathbf{W}}
\newcommand{\matX}{\mathbf{X}}

\newcommand{\tf}{\widetilde{f}}

\newcommand{\setS}{\mathcal{S}}

\newcommand{\bigo}{\mathcal{O}}

\newcommand{\norms}[1]{\|#1\|}

\newcommand{\twonms}[1]{\|#1\|_2}

\newcommand{\pnms}[1]{\|#1\|_p}
\newcommand{\qnm}[1]{\left\|#1\right\|_q}
\newcommand{\qnms}[1]{\|#1\|_q}

\newcommand{\onenms}[1]{\|#1\|_1}

\newcommand{\fbnorms}[1]{\|#1\|_F}

\newcommand{\infnms}[1]{\|#1\|_\infty}

\newcommand{\probs}[1]{\mathbb{P}\{#1\}}

\newcommand{\EE}{\ensuremath{\mathbb{E}}}

\newcommand{\innerps}[2]{\langle#1,#2\rangle}
\newcommand{\abs}[1]{\left|#1\right|}

\newcommand{\B}{\mathcal{B}}
\newcommand{\Ball}{\mathbb{B}}
\newcommand{\D}{\mathcal{D}}

\newcommand{\Z}{\mathcal{Z}}

\newcommand{\A}{\mathcal{A}}
\newcommand{\X}{\mathcal{X}}
\newcommand{\Y}{\mathcal{Y}}
\newcommand{\F}{\mathcal{F}}
\newcommand{\V}{\mathcal{V}}
\newcommand{\calH}{\mathcal{H}}
\newcommand{\vecsigma}{\boldsymbol{\sigma}}
\newcommand{\frakR}{\mathfrak{R}}
\newcommand{\tR}{\widetilde{R}}

\newcommand{\tL}{\widetilde{\ell}}
\newcommand{\tcalF}{\widetilde{\mathcal{F}}}

\algblockdefx{PARFOR}{ENDPARFOR}[1]%
  {\textbf{for all }#1 \textbf{do in parallel}}%
  {\textbf{end for}}

\definecolor{darkgreen}{rgb}{0.0, 0.6, 0.2}
\definecolor{lightgray}{rgb}{0.8, 0.85, 0.85}

\title{Rademacher Complexity for Adversarially Robust Generalization}

\author[1]{Dong Yin \thanks{dongyin@berkeley.edu}}
\author[1]{Kannan Ramchandran \thanks{kannanr@berkeley.edu}}
\author[1,2]{Peter Bartlett \thanks{peter@berkeley.edu}}
\affil[1]{Department of Electrical Engineering and Computer Sciences, UC Berkeley}
\affil[2]{Department of Statistics, UC Berkeley}

\begin{document}
\date{} 
\maketitle

\begin{abstract}
Many machine learning models are vulnerable to adversarial attacks; for example, adding adversarial perturbations that are imperceptible to humans can often make machine learning models produce wrong predictions with high confidence. Moreover, although we may obtain robust models on the training dataset via adversarial training, in some problems the learned models cannot generalize well to the test data. In this paper, we focus on $\ell_\infty$ attacks, and study the adversarially robust generalization problem through the lens of Rademacher complexity. For binary linear classifiers, we prove tight bounds for the adversarial Rademacher complexity, and show that the adversarial Rademacher complexity is never smaller than its natural counterpart, and it has an unavoidable dimension dependence, unless the weight vector has bounded $\ell_1$ norm. The results also extend to multi-class linear classifiers. For (nonlinear) neural networks, we show that the dimension dependence in the adversarial Rademacher complexity also exists. We further consider a surrogate adversarial loss for one-hidden layer ReLU network and prove margin bounds for this setting. Our results indicate that having $\ell_1$ norm constraints on the weight matrices might be a potential way to improve generalization in the adversarial setting. We demonstrate experimental results that validate our theoretical findings. 
\end{abstract} 

\section{Introduction}\label{sec:intro}
In recent years, many modern machine learning models, in particular, deep neural networks, have achieved success in tasks such as image classification~\cite{he2016deep}, speech recognition~\cite{graves2013speech}, machine translation~\cite{bahdanau2014neural}, game playing~\cite{silver2016mastering}, etc. However, although these models achieve the state-of-the-art performance in many standard benchmarks or competitions, it has been observed that by adversarially adding some perturbation to the input of the model (images, audio signals), the machine learning models can make wrong predictions with high confidence. These adversarial inputs are often called the \emph{adversarial examples}. Typical methods of generating adversarial examples include adding small perturbations that are imperceptible to humans~\cite{szegedy2013intriguing}, changing surrounding areas of the main objects in images~\cite{gilmer2018motivating}, and even simple rotation and translation~\cite{engstrom2017rotation}. This phenomenon was first discovered by~\citet{szegedy2013intriguing} in image classification problems, and similar phenomena have been observed in other areas~\cite{carlini2018audio,kos2018adversarial}. Adversarial examples bring serious challenges in many security-critical applications, such as medical diagnosis and autonomous driving---the existence of these examples shows that many state-of-the-art machine learning models are actually unreliable in the presence of adversarial attacks.

Since the discovery of adversarial examples, there has been a race between designing robust models that can defend against adversarial attacks and designing attack algorithms that can generate adversarial examples and fool the machine learning models~\cite{goodfellow6572explaining,gu2014towards,carlini2016defensive,carlini2017adversarial}. As of now, it seems that the attackers are winning this game. For example, a recent work shows that many of the defense algorithms fail when the attacker uses a carefully designed gradient-based method~\cite{athalye2018obfuscated}. Meanwhile, \emph{adversarial training}~\cite{huang2015learning,shaham2015understanding,madry2017towards} seems to be the most effective defense method. Adversarial training takes a robust optimization~\cite{ben2009robust} perspective to the problem, and the basic idea is to minimize some \emph{adversarial loss} over the training data. We elaborate below.

Suppose that data points $(\vecx, y)$ are drawn according to some unknown distribution $\D$ over the feature-label space $\X\times\Y$, and $\X\subseteq \R^d$. Let $\F$ be a hypothesis class (e.g., a class of neural networks with a particular architecture), and $\ell(f(\vecx), y)$ be the loss associated with $f \in \F$. Consider the $\ell_\infty$ white-box adversarial attack where an adversary is allowed to observe the trained model and choose some $\vecx'$ such that $\|\vecx' - \vecx\|_\infty \le \epsilon$ and $\ell(f(\vecx'), y)$ is maximized. Therefore, to better defend against adversarial attacks, during training, the learner should aim to solve the empirical adversarial risk minimization problem
\begin{equation}\label{eq:min_max}
\min_{f\in\F}\frac{1}{n}\sum_{i=1}^n \max_{\| \vecx_i' - \vecx_i \|_\infty \le \epsilon} \ell(f(\vecx_i'), y_i),
\end{equation}
where $\{(\vecx_i, y_i)\}_{i=1}^n$ are $n$ i.i.d. training examples drawn according to $\D$. This minimax formulation raises many interesting theoretical and practical questions. For example, we need to understand how to efficiently solve the optimization problem in~\eqref{eq:min_max}, and in addition, we need to characterize the generalization property of the adversarial risk, i.e., the gap between the empirical adversarial risk in~\eqref{eq:min_max} and the population adversarial risk $\EE_{(\vecx, y)\sim D}[\max_{\| \vecx' - \vecx \|_\infty \le \epsilon} \ell(f(\vecx'), y) ]$. In fact, for deep neural networks, both questions are still wide open. In particular, for the generalization problem, it has been observed that even if we can minimize the adversarial training error, the adversarial test error can still be large. For example, for a Resnet~\cite{he2016deep} model on CIFAR10, using the PGD adversarial training algorithm by~\citet{madry2017towards}, one can achieve about 96\% adversarial training accuracy, but the adversarial test accuracy is only 47\%. This generalization gap is significantly larger than that in the natural setting (without adversarial attacks), and thus it has become increasingly important to better understand the generalization behavior of machine learning models in the adversarial setting.

In this paper, we focus on the adversarially robust generalization property and make a first step towards deeper understanding of this problem. We focus on $\ell_\infty$ adversarial attacks and analyze generalization through the lens of Rademacher complexity. We study both linear classifiers and nonlinear feedforward neural networks, and both binary and multi-class classification problems. We summarize our contributions as follows.

\subsection{Our Contributions}
\begin{itemize}

\item For binary linear classifiers, we prove tight upper and lower bounds for the adversarial Rademacher complexity. We show that the adversarial Rademacher complexity is never smaller than its counterpart in the natural setting, which provides theoretical evidence for the empirical observation that adversarially robust generalization can be hard. We also show that under an $\ell_\infty$ adversarial attack, when the weight vector of the linear classifier has bounded $\ell_p$ norm $(p\ge 1)$, a polynomial dimension dependence in the adversarial Rademacher complexity is unavoidable, unless $p=1$. For multi-class linear classifiers, we prove margin bounds in the adversarial setting. Similar to binary classifiers, the margin bound also exhibits polynomial dimension dependence when the weight vector for each class has bounded $\ell_p$ norm $(p > 1)$.

\item For neural networks, we show that in contrast to the margin bounds derived by~\citet{bartlett2017spectrally} and~\citet{golowich2017size} which depend only on the norms of the weight matrices and the data points, the adversarial Rademacher complexity has a lower bound with an explicit dimension dependence, which is also an effect of the $\ell_\infty$ attack. We further consider a \emph{surrogate adversarial loss} for one hidden layer ReLU networks, based on the SDP relaxation proposed by~\citet{raghunathan2018certified}. We prove margin bounds using the surrogate loss and show that if the weight matrix of the first layer has bounded $\ell_1$ norm, the margin bound does not have explicit dimension dependence. This suggests that in the adversarial setting, controlling the $\ell_1$ norms of the weight matrices may be a way to improve generalization. 

\item We conduct experiments on linear classifiers and neural networks to validate our theoretical findings; more specifically, our experiments show that $\ell_1$ regularization could reduce the adversarial generalization error, and the adversarial generalization gap increases as the dimension of the feature spaces increases.
\end{itemize}

\subsection{Notation} We define the set $[N] := \{1,2,\ldots, N\}$. For two sets $\A$ and $\B$, we denote by $\B^\A$ the set of all functions from $\A$ to $\B$. We denote the indicator function of a event $A$ as $\indi(A)$. Unless otherwise stated, we denote vectors by boldface lowercase letters such as $\vecw$, and the elements in the vector are denoted by italics letters with subscripts, such as $w_k$. All-one vectors are denoted by $ \mathbf{1} $. Matrices are denoted by boldface uppercase letters such as $\matW$. For a matrix $\matW\in\R^{d \times m}$ with columns $\vecw_{i}$, $i\in[m]$, the $(p,q)$ matrix norm of $\matW$ is defined as $\norms{\matW}_{p,q} = \norms{[\norms{\vecw_1}_p,\norms{\vecw_2}_p,\cdots,\norms{\vecw_m}_p ]}_q$, and we may use the shorthand notation $\norms{\cdot}_p \equiv \norms{\cdot}_{p,p}$. We denote the spectral norm of matrices by $\norms{\cdot}_\sigma$ and the Frobenius norm of matrices by $\fbnorms{\cdot}$ (i.e., $\fbnorms{\cdot} \equiv \twonms{\cdot}$). We use $\Ball^\infty_\vecx(\epsilon)$ to denote the $\ell_\infty$ ball centered at $\vecx \in \R^d$ with radius $\epsilon$, i.e., $\Ball^\infty_\vecx(\epsilon) = \{\vecx' \in\R^d : \infnms{\vecx' - \vecx} \le \epsilon\}$. 

\subsection{Organization}
The rest of this paper is organized as follows: in Section~\ref{sec:related}, we discuss related work; in Section~\ref{sec:setup}, we describe the formal problem setup; we present our main results for linear classifiers and neural networks in Sections~\ref{sec:linear_classifiers} and~\ref{sec:neural_nets}, respectively. We demonstrate our experimental results in Section~\ref{sec:experiments} and make conclusions in Section~\ref{sec:conclusions}.

\section{Related Work}\label{sec:related}
During the preparation of the initial draft of this paper, we become aware of another independent and concurrent work by~\citet{khim2018adversarial}, which studies a similar problem. In this section, we first compare our work with~\citet{khim2018adversarial} and then discuss other related work. We make the comparison in the following aspects.
\begin{itemize}[leftmargin=3mm]

\item For binary classification problems, the adversarial Rademacher complexity \emph{upper bound} by~\citet{khim2018adversarial} is similar to ours. However, we provide an adversarial Rademacher complexity \emph{lower bound} that \emph{matches} the upper bound. Our lower bound shows that the adversarial Rademacher complexity is never smaller than that in the natural setting, indicating the hardness of adversarially robust generalization. As mentioned, although our lower bound is for Rademacher complexity rather than generalization, Rademacher complexity is a tight bound for the rate of uniform convergence of a loss function class~\cite{koltchinskii2006local} and thus in many settings can be a tight bound for generalization. In addition, we provide a lower bound for the adversarial Rademacher complexity for neural networks. These lower bounds do not appear in the work by~\citet{khim2018adversarial}.

\item We discuss the generalization bounds for the multi-class setting, whereas~\citet{khim2018adversarial} focus only on binary classification.

\item Both our work and~\citet{khim2018adversarial} prove adversarial generalization bound using surrogate adversarial loss (upper bound for the actual adversarial loss). \citet{khim2018adversarial} use a method called \emph{tree transform} whereas we use the SDP relaxation proposed by~\cite{raghunathan2018certified}. These two approaches are based on different ideas and thus we believe that they are not directly comparable.

\end{itemize}

We proceed to discuss other related work.

\paragraph*{Adversarially robust generalization} As discussed in Section~\ref{sec:intro}, it has been observed by~\citet{madry2017towards} that there might be a significant generalization gap when training deep neural networks in the adversarial setting. This generalization problem has been further studied by~\citet{schmidt2018adversarially}, who show that to correctly classify two separated $d$-dimensional spherical Gaussian distributions, in the natural setting one only needs $\bigo(1)$ training data, but in the adversarial setting one needs $\Theta(\sqrt{d})$ data. Getting distribution agnostic generalization bounds (also known as the PAC-learning framework) for the adversarial setting is proposed as an open problem by~\citet{schmidt2018adversarially}. In a subsequent work,~\citet{cullina2018pac} study PAC-learning guarantees for binary linear classifiers in the adversarial setting via VC-dimension, and show that the VC-dimension does not increase in the adversarial setting. This result does not provide explanation to the empirical observation that adversarially robust generalization may be hard. In fact, although VC-dimension and Rademacher complexity can both provide valid generalization bounds, VC-dimension usually depends on the number of parameters in the model while Rademacher complexity usually depends on the norms of the weight matrices and data points, and can often provide tighter generalization bounds~\cite{b-scpcnn-98}. \citet{suggala2018adversarial} discuss a similar notion of adversarial risk but do not prove explicit generalization bounds. \citet{attias2018improved} prove adversarial generalization bounds in a setting where the number of potential adversarial perturbations is finite, which is a weaker notion than the $\ell_\infty$ attack that we consider.~\citet{sinha2018certifying} analyze the convergence and generalization of an adversarial training algorithm under the notion of distributional robustness.~\citet{farnia2018generalizable} study the generalization problem when the attack algorithm of the adversary is provided to the learner, which is also a weaker notion than our problem. In earlier work, robust optimization has been studied in Lasso~\cite{xu2009robust} and SVM~\cite{xu2009robustness} problems.~\citet{xu2012robustness} make the connection between algorithmic robustness and generalization property in the natural setting, whereas our work focus on generalization in the adversarial setting.

\paragraph*{Provable defense against adversarial attacks} Besides generalization property, another recent line of work aim to design provable defense against adversarial attacks. Two examples of provable defense are SDP relaxation~\cite{raghunathan2018certified,raghunathan2018semidefinite} and LP relaxation~\cite{kolter2017provable,wong2018scaling}. The idea of these methods is to construct \emph{upper bounds} of the adversarial risk that can be efficiently evaluated and optimized. The analyses of these algorithms usually focus on minimizing training error and do not have generalization guarantee; in contrast, we focus on generalization property in this paper. 

\paragraph*{Other theoretical analysis of adversarial examples} A few other lines of work have conducted theoretical analysis of adversarial examples.~\citet{wang2017analyzing} analyze the adversarial robustness of nearest neighbors estimator.~\citet{papernot2016towards} try to demonstrate the unavoidable trade-offs between accuracy in the natural setting and the resilience to adversarial attacks, and this trade-off is further studied by~\citet{tsipras2018there} through some constructive examples of distributions.~\citet{fawzi2016robustness} analyze adversarial robustness of fixed classifiers, in contrast to our generalization analysis.~\citet{fawzi2018adversarial} construct examples of distributions with large latent variable space such that adversarially robust classifiers do not exist; here we argue that these examples may not explain the fact that adversarially perturbed images can usually be recognized by humans.~\citet{bubeck2018adversarial} try to explain the hardness of learning an adversarially robust model from the computational constraints under the statistical query model. Another recent line of work explains the existence of adversarial examples via high dimensional geometry and concentration of measure~\cite{gilmer2018adversarial,dohmatob2018limitations,mahloujifar2018curse}. These works provide examples where adversarial examples provably exist as long as the test error of a classifier is non-zero.

\paragraph*{Generalization of neural networks} Generalization of neural networks has been an important topic, even in the natural setting where there is no adversary. The key challenge is to understand why deep neural networks can generalize to unseen data despite the high capacity of the model class. The problem has received attention since the early stage of neural network study~\cite{b-scpcnn-98,anthony2009neural}. Recently, understanding generalization of deep nets is raised as an open problem since traditional techniques such as VC-dimension, Rademacher complexity, and algorithmic stability are observed to produce vacuous generalization bounds~\cite{zhang2016understanding}. Progress has been made more recently. In particular, it has been shown that when properly normalized by the margin, using Rademacher complexity or PAC-Bayesian analysis, one can obtain generalization bounds that tend to match the experimental results~\cite{bartlett2017spectrally,neyshabur2017pac,arora2018stronger,golowich2017size}. In addition, in this paper, we show that when the weight vectors or matrices have bounded $\ell_1$ norm, there is no dimension dependence in the adversarial generalization bound. This result is consistent and related to several previous works~\cite{lee1996efficient,b-scpcnn-98,mei2018mean,zhang2016l1}.

\section{Problem Setup}\label{sec:setup} 

We start with the general statistical learning framework. Let $\X$ and $\Y$ be the feature and label spaces, respectively, and suppose that there is an unknown distribution $\D$ over $\X \times \Y$. In this paper, we assume that the feature space is a subset of the $d$ dimensional Euclidean space, i.e., $\X \subseteq \R^d$. Let $\F \subseteq \V^\X$ be the hypothesis class that we use to make predictions, where $\V$ is another space that might be different from $\Y$. Let $\ell : \V \times \Y \rightarrow [0, B]$ be the loss function. Throughout this paper we assume that $\ell$ is bounded, i.e., $B$ is a positive constant. In addition, we introduce the function class $\ell_\F\subseteq [0,B]^{\X \times \Y}$ by composing the functions in $\F$ with $\ell(\cdot, y)$, i.e., $\ell_\F := \{ (\vecx, y) \mapsto \ell(f(\vecx), y) : f\in\F\}$. The goal of the learning problem is to find $f \in \F$ such that the \emph{population risk} $R(f) := \EE_{(\vecx, y)\in\D} [\ell(f(\vecx), y)]$ is minimized.

We consider the supervised learning setting where one has access to $n$ i.i.d. training examples drawn according to $\D$, denoted by $(\vecx_1, y_1), (\vecx_2, y_2), \ldots, (\vecx_n, y_n)$. A learning algorithm maps the $n$ training examples to a hypothesis $f\in \F$. In this paper, we are interested in the gap between the \emph{empirical risk} $R_n(f) := \frac{1}{n} \sum_{i=1}^n \ell(f(\vecx_i), y_i)$ and the population risk $R(f)$, known as the generalization error.

Rademacher complexity~\cite{bartlett2002rademacher} is one of the classic measures of generalization error. Here, we present its formal definition. For any function class $\calH \subseteq \R^{\Z}$, given a sample $\setS=\{\vecz_1, \vecz_2, \ldots, \vecz_n\}$ of size $n$, and \emph{empirical Rademacher complexity} is defined as 
\[
\frakR_{\setS}(\calH) := \frac{1}{n}\EE_{\vecsigma} \left[ \sup_{h \in \calH} \sum_{i=1}^n \sigma_ih(\vecz_i) \right], 
\]
where $\sigma_1,\ldots,\sigma_n$ are i.i.d. Rademacher random variables with $\probs{\sigma_i=1} = \probs{\sigma_i=-1} = \frac{1}{2}$. In our learning problem, denote the training sample by $\setS$, i.e., $\setS := \{(\vecx_1, y_1), (\vecx_2, y_2), \ldots, (\vecx_n, y_n)\}$. We then have the following theorem which connects the population and empirical risks via Rademacher complexity.
\begin{theorem}\label{thm:rad_vanilla}
\cite{bartlett2002rademacher,mohri2012foundations} Suppose that the range of $\ell(f(\vecx), y)$ is $[0,B]$. Then, for any $\delta \in (0,1)$, with probability at least $1-\delta$, the following holds for all $f\in\F$,
\[
R(f) \le R_n(f) + 2B\frakR_{\setS}(\ell_{\F}) + 3B\sqrt{\frac{\log \frac{2}{\delta}}{2n}}.
\]
\end{theorem}
As we can see, Rademacher complexity measures the rate that the empirical risk converges to the population risk \emph{uniformly} across $\F$. In fact, according to the anti-symmetrization lower bound by~\citet{koltchinskii2006local}, one can show that $\frakR_{\setS}(\ell_{\F}) \lesssim \sup_{f\in\F} R(f) - R_n(f) \lesssim \frakR_{\setS}(\ell_{\F})$. Therefore, Rademacher complexity is a tight bound for the rate of uniform convergence of a loss function class, and in many settings can be a tight bound for generalization error. 

The above discussions can be extended to the adversarial setting. In this paper, we focus on the $\ell_\infty$ adversarial attack. In this setting, the learning algorithm still has access to $n$ i.i.d. uncorrupted training examples drawn according to $\D$. Once the learning procedure finishes, the output hypothesis $f$ is revealed to an adversary. For any data point $(\vecx, y)$ drawn according to $\D$, the adversary is allowed to perturb $\vecx$ within some $\ell_\infty$ ball to maximize the loss. Our goal is to minimize the \emph{adversarial population risk}, i.e.,
\[
\tR(f) := \EE_{(\vecx,y)\sim\D} \left[ \max_{  \vecx' \in \Ball^\infty_\vecx(\epsilon)  }\ell(f(\vecx'), y) \right],
\]
and to this end, a natural way is to conduct \emph{adversarial training}---minimizing the adversarial empirical risk
\[
\tR_n(f) := \frac{1}{n}\sum_{i=1}^n \max_{ \vecx_i' \in \Ball^\infty_{\vecx_i}(\epsilon) } \ell(f(\vecx_i'), y_i).
\] 
Let us define the adversarial loss $\tL(f(\vecx), y) := \max_{ \Ball^\infty_\vecx(\epsilon) } \ell(f(\vecx'), y)$ and the function class $\tL_{\F} \subseteq [0,B]^{\X \times \Y} $ as $
\tL_{\F} := \{\tL(f(\vecx), y) : f\in\F\}$.
Since the range of $\tL(f(\vecx), y)$ is still $[0, B]$, we have the following direct corollary of Theorem~\ref{thm:rad_vanilla}.
\begin{cor}\label{cor:rad_gen}
For any $\delta \in (0,1)$, with probability at least $1-\delta$, the following holds for all $f\in\F$,
\[
\tR(f) \le \tR_n(f) + 2B\frakR_{\setS}(\tL_{\F}) + 3B\sqrt{\frac{\log \frac{2}{\delta}}{2n}}.
\]
\end{cor}
As we can see, the Rademacher complexity of the adversarial loss function class $\tL_{\F}$, i.e., $\frakR_{\setS}(\tL_{\F})$ is again the key quantity for the generalization ability of the learning problem. A natural problem of interest is to compare the Rademacher complexities in the natural and the adversarial settings, and obtain generalization bounds for the adversarial loss.

\section{Linear Classifiers}\label{sec:linear_classifiers}

\subsection{Binary Classification}\label{sec:binary_classification}

We start with binary linear classifiers. In this setting, we define $\Y = \{-1, +1\}$, and let the hypothesis class $\F\subseteq \R^\X$ be a set of linear functions of $\vecx\in\X$. More specifically, we define $f_\vecw(\vecx) := \innerps{\vecw}{\vecx}$, and consider prediction vector $\vecw$ with $\ell_p$ norm constraint ($p\ge 1$), i.e., 

\begin{equation}\label{eq:def_binary_nat}
\F = \{ f_\vecw(\vecx) : \pnms{\vecw} \le W \}.
\end{equation}

We predict the label with the sign of $f_\vecw(\vecx)$; more specifically, we assume that the loss function $\ell(f_\vecw(\vecx), y)$ can be written as $\ell(f_\vecw(\vecx), y) \equiv \phi(y\innerps{\vecw}{\vecx})$, where $\phi : \R\rightarrow [0,B]$ is monotonically nonincreasing and $L_\phi$-Lipschitz. In fact, if $\phi(0) \ge 1$, we can obtain a bound on the classification error according to Theorem~\ref{thm:rad_vanilla}. That is, with probability at least $1-\delta$, for all $f_\vecw \in\F$,
\[
\PP_{(\vecx, y)\sim\D}\{\sgn(f_\vecw(\vecx)) \neq y\} \le \frac{1}{n}\sum_{i=1}^n \ell(f_\vecw(\vecx_i), y_i) + 2B\frakR_{\setS}(\ell_{\F}) + 3B\sqrt{\frac{\log \frac{2}{\delta}}{2n}}.
\]
In addition, recall that according to the Ledoux-Talagrand contraction inequality~\cite{ledoux2013probability}, we have 
$\frakR_{\setS}(\ell_{\F}) \le L_\phi \frakR_{\setS}(\F)$.

For the adversarial setting, we have 
\[
\tL(f_\vecw(\vecx), y) \hspace{-0.3mm} = \hspace{-0.3mm} \hspace{-1.2mm} \max_{ \vecx' \in \Ball^\infty_\vecx(\epsilon) } \hspace{-0.8mm} \ell(f_\vecw(\vecx'), y) 
\hspace{-0.6mm} = \hspace{-0.6mm} \phi( \hspace{-0.8mm} \min_{ \vecx' \in \Ball^\infty_\vecx(\epsilon) } \hspace{-0.8mm} y\innerps{\vecw}{ \vecx' }).
\]
Let us define the following function class $\tcalF \subseteq \R^{\X \times \{\pm1\}}$:

\begin{equation}\label{eq:def_binary_adv}
\tcalF = \left\{ \min_{ \vecx' \in \Ball^\infty_\vecx(\epsilon) } y\innerps{\vecw}{\vecx'} :  \pnms{\vecw} \le W \right\}.
\end{equation}
Again, we have $\frakR_{\setS}(\tL_{\F}) \le L_\phi \frakR_{\setS}(\tcalF)$. The first major contribution of our work is the following theorem, which provides a comparison between $\frakR_{\setS}(\F)$ and $\frakR_{\setS}(\tcalF)$.
\begin{theorem}[\textbf{Main Result 1}]\label{thm:rad_compare}
Let $\F := \{ f_\vecw(\vecx) : \pnms{\vecw} \le W \}$ and $\tcalF := \{ \min_{ \vecx' \in \Ball^\infty_\vecx(\epsilon) } y\innerps{\vecw}{\vecx'} :  \pnms{\vecw} \le W \}$. Suppose that $\frac{1}{p} + \frac{1}{q} = 1$. Then, there exists a universal constant $c \in (0, 1)$ such that 
\[
\max\{\frakR_{\setS}(\F), c\epsilon W \frac{d^{\frac{1}{q}}}{\sqrt{n}} \} \le \frakR_{\setS}(\tcalF) \le \frakR_{\setS}(\F) + \epsilon W \frac{d^{\frac{1}{q}}}{\sqrt{n}}.
\]
\end{theorem} 

We prove Theorem~\ref{thm:rad_compare} in Appendix~\ref{prf:rad_compare}. We can see that the adversarial Rademacher complexity, i.e., $\frakR_{\setS}(\tcalF)$ is always at least as large as the Rademacher complexity in the natural setting. This implies that uniform convergence in the adversarial setting is at least as hard as that in the natural setting. In addition, since $\max\{a,b\} \ge \frac{1}{2}(a+b)$, we have
\[
\frac{c}{2} \Big( \frakR_{\setS}(\F) + \epsilon W \frac{d^{\frac{1}{q}}}{\sqrt{n}}  \Big) \le \frakR_{\setS}(\tcalF) \le \frakR_{\setS}(\F) + \epsilon W \frac{d^{\frac{1}{q}}}{\sqrt{n}}.
\]
Therefore, we have a tight bound for $\frakR_{\setS}(\tcalF)$ up to a constant factor. Further, if $p>1$ the adversarial Rademacher complexity has an unavoidable polynomial dimension dependence, i.e., $\frakR_{\setS}(\tcalF)$ is always at least as large as $\bigo(\epsilon W \frac{d^{1/q}}{\sqrt{n}})$. On the other hand, one can easily show that in the natural setting, $\frakR_{\setS}(\F)= \frac{W}{n} \EE_{\vecsigma} [ \qnms{\sum_{i=1}^n \sigma_i \vecx_i} ]$, which implies that $\frakR_{\setS}(\F)$ depends on the distribution of $\vecx_i$ and the norm constraint $W$, but does not have an explicit dimension dependence. This means that $\frakR_{\setS}(\tcalF)$ could be order-wise larger than $\frakR_{\setS}(\F)$, depending on the distribution of the data. An interesting fact is that, if we have an $\ell_1$ norm constraint on the prediction vector $\vecw$, we can avoid the dimension dependence in $\frakR_{\setS}(\tcalF)$.

\subsection{Multi-class Classification}

\subsubsection{Margin Bounds for Multi-class Classification}

We proceed to study multi-class linear classifiers. We start with the standard margin bound framework for multi-class classification. In $K$-class classification problems, we choose $\Y = [K]$, and the functions in the hypothesis class $\F$ map $\X$ to $\R^K$, i.e., $\F \subseteq (\R^K)^\X$. Intuitively, the $k$-th coordinate of $f(\vecx)$ is the score that $f$ gives to the $k$-th class, and we make prediction by choosing the class with the highest score. We define the margin operator $M(\vecz, y) : \R^K \times [K] \rightarrow \R$ as $M(\vecz, y) = z_y - \max_{y' \neq y} z_{y'}$. For a training example $(\vecx, y)$, a hypothesis $f$ makes a correct prediction if and only if $M(f(\vecx), y) > 0$. We also define function class $M_\F := \{ (\vecx, y) \mapsto M(f(\vecx), y) : f \in \F \}\subseteq \R^{\X \times [K]}$. For multi-class classification problems, we consider a particular loss function $\ell (f(\vecx), y) = \phi_\gamma (M(f(\vecx), y))$, where $\gamma > 0$ and $\phi_\gamma:\R \rightarrow [0,1]$ is the ramp loss:
\begin{equation}\label{eq:ramp}
\phi_\gamma(t) = \begin{cases}
1 & t \le 0 \\
1-\frac{t}{\gamma} & 0 < t < \gamma \\
0 & t \ge \gamma.
\end{cases}
\end{equation}
One can check that $\ell (f(\vecx), y)$ satisfies:
\begin{equation}\label{eq:ramp_property}
\indi(y \neq \arg\max_{y' \in [K]} [f(\vecx)]_{y'}) \le \ell(f(\vecx), y) \le \indi([f(\vecx)]_y \le \gamma + \max_{y' \neq y}[f(\vecx)]_{y'}).
\end{equation}
Let $\setS = \{(\vecx_i, y_i)\}_{i=1}^n \in (\X \times [K])^n$ be the i.i.d. training examples, and define the function class $\ell_{\F} := \{(\vecx, y) \mapsto \phi_\gamma (M(f(\vecx), y)) : f\in\F\} \subseteq \R^{\X\times [K]}$. Since $\phi_\gamma(t) \in [0,1]$ and $\phi_\gamma(\cdot)$ is $1/\gamma$-Lipschitz, by combining~\eqref{eq:ramp_property} with Theorem~\ref{thm:rad_vanilla}, we can obtain the following direct corollary as the generalization bound in the multi-class setting~\cite{mohri2012foundations}.
\begin{cor}\label{cor:multi_natural}
Consider the above multi-class classification setting. For any fixed $\gamma >0$, we have with probability at least $1-\delta$, for all $f \in \F$, 
\[
\PP_{(\vecx, y)\sim\D} \left\{ y \neq \arg\max_{y' \in [K]} [f(\vecx)]_{y'} \right\} \le \frac{1}{n}\sum_{i=1}^n \indi([f(\vecx_i)]_{y_i} \le \gamma + \max_{y' \neq y}[f(\vecx_i)]_{y'}) + 2\frakR_{\setS}(\ell_{\F}) + 3\sqrt{\frac{\log \frac{2}{\delta}}{2n}}.
\]
\end{cor}

In the adversarial setting, the adversary tries to maximize the loss $\ell (f(\vecx), y) = \phi_\gamma (M(f(\vecx), y))$ around an $\ell_\infty$ ball centered at $\vecx$. We have the adversarial loss $\tL(f(\vecx), y) := \max_{ \vecx' \in \Ball^\infty_\vecx(\epsilon) } \ell(f(\vecx'), y)$, and the function class $\tL_{\F} := \{(\vecx, y) \mapsto \tL(f(\vecx), y) : f\in\F\} \subseteq \R^{\X\times [K]}$. Thus, we have the following generalization bound in the adversarial setting.

\begin{cor}\label{cor:multi_adv}
Consider the above adversarial multi-class classification setting. For any fixed $\gamma >0$, we have with probability at least $1-\delta$, for all $f \in \F$,
\begin{align*}
& \PP_{(\vecx, y)\sim\D} \left\{ \exists~\vecx' \in \Ball^\infty_\vecx(\epsilon)~\text{s.t.}~ y \neq \arg\max_{y' \in [K]} [f(\vecx')]_{y'} \right\} \\
\le & \frac{1}{n}\sum_{i=1}^n \indi( \exists~\vecx_i' \in \Ball^\infty_{\vecx_i} (\epsilon)~\text{s.t.}~ [f( \vecx_i' )]_{y_i} \le \gamma + \max_{y' \neq y}[f( \vecx_i' )]_{y'}) + 2\frakR_{\setS}(\tL_{\F}) + 3\sqrt{\frac{\log \frac{2}{\delta}}{2n}}.
\end{align*}
\end{cor}

\subsubsection{Multi-class Linear Classifiers} 

We now focus on multi-class linear classifiers. For linear classifiers, each function in the hypothesis class is linearly parametrized by a matrix $\matW\in\R^{K\times d}$, i.e., $f(\vecx)\equiv f_{\matW}(\vecx) = \matW \vecx$. Let $\vecw_k \in \R^d$ be the $k$-th column of $\matW^\top$, then we have $[f_\matW(\vecx)]_k = \innerps{\vecw_k}{\vecx}$. We assume that each $\vecw_k$ has $\ell_p$ norm ($p \ge 1$) upper bounded by $W$, which implies that $\F = \{ f_\matW(\vecx) : \norms{\matW^\top}_{p,\infty} \le W \}$. In the natural setting, we have the following margin bound for linear classifiers as a corollary of the multi-class margin bounds by~\citet{kuznetsov2015rademacher,maximov2016tight}.
\begin{theorem}\label{thm:multi_nat}
Consider the multi-class linear classifiers in the above setting, and suppose that $\frac{1}{p} + \frac{1}{q} = 1$, $p,q \ge 1$. For any fixed $\gamma >0$ and $W > 0$, we have with probability at least $1-\delta$, for all $\matW$ such that $\norms{\matW^\top}_{p,\infty} \le W$,
\begin{align*}
 \PP_{(\vecx, y)\sim\D} \left\{ y \neq \arg\max_{y' \in [K]} \innerps{\vecw_{y'}}{\vecx} \right\}  \le & \frac{1}{n}\sum_{i=1}^n \indi(\innerps{\vecw_{y_i}}{\vecx_i} \le \gamma + \max_{y' \neq y_i} \innerps{\vecw_{y'}}{\vecx_i} ) \\
 & + \frac{4KW}{\gamma n }\EE_{\vecsigma} \left[ \qnms{\sum_{i=1}^n \sigma_i\vecx_i} \right] + 3\sqrt{\frac{\log \frac{2}{\delta}}{2n}}.
\end{align*}
\end{theorem}
We prove Theorem~\ref{thm:multi_nat} in Appendix~\ref{prf:multi_nat} for completeness. In the adversarial setting, we have the following margin bound.
\begin{theorem}[\textbf{Main Result 2}]\label{thm:multi_adv}
Consider the multi-class linear classifiers in the adversarial setting, and suppose that $\frac{1}{p} + \frac{1}{q} = 1$, $p,q \ge 1$. For any fixed $\gamma >0$ and $W > 0$, we have with probability at least $1-\delta$, for all $\matW$ such that $\norms{\matW^\top}_{p,\infty} \le W$,\begin{align*}
& \PP_{(\vecx, y)\sim\D} \left\{ \exists~\vecx' \in \Ball^\infty_\vecx(\epsilon)~\text{s.t.}~y \neq \arg\max_{y' \in [K]} \innerps{\vecw_{y'}}{\vecx'} \right\}  \\
\le & \frac{1}{n}\sum_{i=1}^n \indi(\innerps{\vecw_{y_i}}{\vecx_i} \le \gamma + \max_{y' \neq y_i} (\innerps{\vecw_{y'}}{\vecx_i} + \epsilon \onenms{\vecw_{y'} - \vecw_{y_i}}) )  \\
& + \frac{2WK}{\gamma} \left[ \frac{\epsilon\sqrt{K} d^{\frac{1}{q}}}{\sqrt{n}} + \frac{1}{n} \sum_{y=1}^K \EE_{\vecsigma} \Big[ \qnms{ \sum_{i=1}^n \sigma_i \vecx_i\indi(y_i = y)} \Big]  \right] + 3\sqrt{\frac{\log \frac{2}{\delta}}{2n}}.
\end{align*}

\end{theorem}
We prove Theorem~\ref{thm:multi_adv} in Appendix~\ref{prf:multi_adv}. As we can see, similar to the binary classification problems, if $ p > 1$, the margin bound in the adversarial setting has an explicit polynomial dependence on $ d $, whereas in the natural setting, the margin bound does not have dimension dependence. This shows that, at least for the generalization upper bound that we obtain, the dimension dependence in the adversarial setting also exists in the multi-class classification problems.

\section{Neural Networks}\label{sec:neural_nets} 

We proceed to consider feedforward neural networks with ReLU activation. Here, each function $f$ in the hypothesis class $\F$ is parametrized by a sequence of matrices $\matW = (\matW_1, \matW_2, \ldots, \matW_L)$, i.e., $f \equiv f_\matW$. Assume that $\matW_h \in \R^{d_h \times d_{h-1}}$, and $\rho(\cdot)$ be the ReLU function, i.e., $\rho(t) = \max\{t, 0\}$ for $t\in\R$. For vectors, $\rho(\vecx)$ is vector generated by applying $\rho(\cdot)$ on each coordinate of $\vecx$, i.e., $[ \rho(\vecx) ]_i = \rho(x_i)$. We have\footnote{This implies that $d_0 \equiv d$.}
\[
f_\matW (\vecx) = \matW_L \rho ( \matW_{L-1} \rho (\cdots  \rho(\matW_{1}\vecx)  \cdots)  ).
\]
For $K$-class classification, we have $d_L = K$, $f_\matW (\vecx) : \R^d \rightarrow \R^K$, and $[f_\matW (\vecx)]_k$ is the score for the $k$-th class. In the special case of binary classification, as discussed in Section~\ref{sec:binary_classification}, we can have $\Y = \{+1, -1\}$, $d_L = 1$, and the loss function can be written as
\[
\ell(f_\matW(\vecx), y) = \phi(yf_\matW(\vecx)),
\]
where $\phi : \R \rightarrow [0, B]$ is monotonically nonincreasing and $L_\phi$-Lipschitz.

\subsection{Comparison of Rademacher Complexity Bounds}
We start with a comparison of Rademacher complexities of neural networks in the natural and adversarial settings. Although naively applying the definition of Rademacher complexity may provide a loose generalization bound~\cite{zhang2016understanding}, when properly normalized by the margin, one can still derive generalization bound that matches experimental observations via Rademacher complexity~\cite{bartlett2017spectrally}. Our comparison shows that, when the weight matrices of the neural networks have bounded norms, in the natural setting, the Rademacher complexity is upper bounded by a quantity which only has logarithmic dependence on the dimension; however, in the adversarial setting, the Rademacher complexity is lower bounded by a quantity with explicit $\sqrt{d}$ dependence.

We focus on the binary classification. For the natural setting, we review the results by~\citet{bartlett2017spectrally}. Let $\setS = \{(\vecx_i, y_i)\}_{i=1}^n \in (\X \times \{-1, +1\})^n$ be the i.i.d. training examples, and define $\matX := [\vecx_1,\vecx_2, \cdots, \vecx_n]\in\R^{d \times n}$, and $d_{\max} = \max\{d, d_1,d_2,\ldots, d_L\}$.

\begin{theorem}\label{thm:nn_natural}
~\cite{bartlett2017spectrally} Consider the neural network hypothesis class
\[
\F = \{f_\matW(\vecx) : \matW = (\matW_1,\matW_2,\ldots, \matW_L), \| \matW_h \|_\sigma \le s_h,  \| \matW_h^\top \|_{2,1}\le b_h, h\in[L] \} \subseteq \R^\X.
\]
Then, we have
\[
\frakR_{\setS}(\F) \le \frac{4}{n^{3/2}} + \frac{26\log(n) \log(2d_{\max})}{n}\fbnorms{\matX} \left( \prod_{h=1}^L s_h \right) \left( \sum_{j=1}^L (\frac{b_j}{s_j})^{2/3} \right)^{3/2}.
\]
\end{theorem}
On the other hand, in this work, we prove the following result which shows that when the product of the spectral norms of all the weight matrices is bounded, the Rademacher complexity of the adversarial loss function class is lower bounded by a quantity with an explicit $\sqrt{d}$ factor. More specifically, for binary classification problems, since 
\[
\tL(f_\matW(\vecx), y) = \max_{ \vecx' \in \Ball^\infty_\vecx(\epsilon) } \ell(f_\matW(\vecx'), y) = \phi(\min_{ \vecx' \in \Ball^\infty_\vecx(\epsilon) } yf_\matW(\vecx')),
\]
and $\phi(\cdot)$ is Lipschitz, we consider the function class
\begin{equation}\label{eq:def_adv_f}
\tcalF = \{(\vecx, y) \mapsto \min_{ \vecx' \in \Ball^\infty_\vecx(\epsilon) } y f_\matW(\vecx') : \matW = (\matW_1,\matW_2,\ldots, \matW_L), \prod_{h=1}^L \| \matW_h \|_\sigma \le r \}  \subseteq \R^{\X \times \{-1, +1\}}.
\end{equation}
Then we have the following result. 
\begin{theorem}[\textbf{Main Result 3}]\label{thm:nn_adv}
Let $\tcalF$ be defined as in~\eqref{eq:def_adv_f}. Then, there exists a universal constant $c>0$ such that
\[
\frakR_\setS(\tcalF) \ge cr \left(\frac{1}{n} \fbnorms{\matX} + \epsilon\sqrt{\frac{d}{n}} \right).
\]
\end{theorem}
We prove Theorem~\ref{thm:nn_adv} in Appendix~\ref{prf:nn_adv}. This result shows that if we aim to study the Rademacher complexity of the function class defined as in~\eqref{eq:def_adv_f}, a $\sqrt{d}$ dimension dependence may be unavoidable, in contrast to the natural setting where the dimension dependence is only logarithmic.

\subsection{Generalization Bound for Surrogate Adversarial Loss}
For neural networks, even if there is only one hidden layer, for a particular data point $(\vecx, y)$, evaluating the adversarial loss $\tL(f_\matW(\vecx), y) = \max_{ \vecx' \in \Ball^\infty_\vecx(\epsilon) } \ell(f_\matW(\vecx'), y)$ can be hard, since it requires maximizing a non-concave function in a bounded set. A recent line of work tries to find upper bounds for $\tL(f_\matW(\vecx), y)$ that can be computed in polynomial time. More specifically, we would like to find \emph{surrogate adversarial loss} $\widehat{\ell}(f_\matW(\vecx), y)$ such that $ \widehat{\ell}(f_\matW(\vecx), y) \ge \tL(f_\matW(\vecx), y)$, $\forall~\vecx,y,\matW$. These surrogate adversarial loss can thus provide \emph{certified} defense against adversarial examples, and can be computed efficiently. In addition, the surrogate adversarial loss $\widehat{\ell}(f_\matW(\vecx), y)$ should be as tight as possible---it should be close enough to the original adversarial loss $\tL(f_\matW(\vecx), y)$, so that the surrogate adversarial loss can indeed represent the robustness of the model against adversarial attacks. Recently, a few approaches to designing surrogate adversarial loss have been developed, and SDP relaxation~\cite{raghunathan2018certified,raghunathan2018semidefinite} and LP relaxation~\cite{kolter2017provable,wong2018scaling} are two major examples.

In this section, we focus on the SDP relaxation for one hidden layer neural network with ReLU activation proposed by~\citet{raghunathan2018certified}. We prove a generalization bound regarding the surrogate adversarial loss, and show that this generalization bound does not have explicit dimension dependence if the weight matrix of the first layer has bounded $\ell_1$ norm. We consider $K$-class classification problems in this section (i.e., $\Y = [K]$), and start with the definition and property of the SDP surrogate loss. Since we only have one hidden layer, $f_\matW(\vecx) = \matW_2\rho(\matW_1\vecx)$. Let $\vecw_{2,k}$ be the $k$-th column of $\matW_2^\top$. Then, we have the following results according to~\citet{raghunathan2018certified}.
\begin{theorem}\label{thm:sdp_surrogate}
~\cite{raghunathan2018certified} For any $(\vecx, y)$, $\matW_1$, $\matW_2$, and $y' \neq y$,
\[
\max_{\vecx' \in \Ball^\infty_\vecx(\epsilon)} ([f_\matW(\vecx')]_{y'}  -  [f_\matW(\vecx')]_{y}) \le [f_\matW(\vecx)]_{y'}  -  [f_\matW(\vecx)]_{y} + \frac{\epsilon}{4} \max_{\matP\succeq 0, \diag(\matP) \le 1} \innerps{Q(\vecw_{2,y'} - \vecw_{2,y}, \matW_1)}{\matP},
\]
where $Q(\vecv, \matW)$ is defined as
\begin{equation}\label{eq:def_Q}
Q(\vecv, \matW) := \left[ \begin{array}{ccc}
0 & 0 & \mathbf{1}^\top \matW^\top \diag(\vecv) \\
0 & 0 & \matW^\top \diag(\vecv)  \\
\diag(\vecv)^\top \matW  \mathbf{1}  & \diag(\vecv)^\top \matW & 0
\end{array} \right].
\end{equation}
\end{theorem}
Since we consider multi-class classification problems in this section, we use the ramp loss $\phi_\gamma$ defined in~\eqref{eq:ramp} composed with the margin operator as our loss function. Thus, we have $\ell(f_\matW(\vecx), y) = \phi_\gamma (M(f_\matW(\vecx), y))$ and $\tL(f_\matW(\vecx), y) = \max_{  \vecx' \in \Ball^\infty_\vecx(\epsilon)  } \phi_\gamma (M(f_\matW(\vecx'), y))$. Here, we design a surrogate loss $\widehat{\ell }(f_\matW(\vecx), y)$ based on Theorem~\ref{thm:sdp_surrogate}. 

\begin{lemma}\label{lem:surrogate}
Define
\[
\widehat{\ell}(f_\matW(\vecx), y) := \phi_\gamma \Big( M(f_\matW(\vecx) , y) - \frac{\epsilon}{2} \max_{k\in[K] , z=\pm1} \max_{\matP\succeq 0, \diag(\matP) \le 1} \innerps{zQ(\vecw_{2,k}, \matW_1)}{\matP}  \Big).
\]
Then, we have 
\begin{align*}
& \max_{\vecx' \in \Ball^\infty_\vecx(\epsilon)} \indi(y \neq \arg\max_{y'\in[K]} [f_\matW(\vecx')]_{y'} ) \le \widehat{\ell}(f_\matW(\vecx), y)  \\
& \le \indi\Big( M(f_\matW(\vecx) , y) - \frac{\epsilon}{2} \max_{k\in[K] , z=\pm1} \max_{\matP\succeq 0, \diag(\matP) \le 1} \innerps{zQ(\vecw_{2,k}, \matW_1)}{\matP}  \le \gamma \Big).
\end{align*}
\end{lemma}
We prove Lemma~\ref{lem:surrogate} in Appendix~\ref{prf:surrogate}. With this surrogate adversarial loss in hand, we can develop the following margin bound for adversarial generalization. In this theorem, we use $\matX = [\vecx_1,\vecx_2, \cdots, \vecx_n]\in\R^{d \times n}$, and $d_{\max} = \max\{d, d_1,K\}$.
\begin{theorem}\label{thm:sdp_gen}
Consider the neural network hypothesis class  
\[
\F = \{ f_\matW(\vecx) : \matW = (\matW_1, \matW_2), \| \matW_h \|_\sigma \le s_h, h=1,2, \|\matW_1 \|_{1} \le b_1,  \|\matW_2^\top \|_{2,1} \le b_2\}.
\]
Then, for any fixed $\gamma > 0$, with probability at least $1-\delta$, we have for all $f_\matW(\cdot)\in\F$,
\begin{align*}
& \PP_{(\vecx, y)\sim\D} \left\{ \exists~\vecx' \in \Ball^\infty_\vecx(\epsilon)~\text{s.t.}~y \neq \arg\max_{y' \in [K]}  [f_\matW(\vecx')]_{y'} \right\}  \\
\le & \frac{1}{n} \sum_{i=1}^n \indi\Big( [f_\matW(\vecx_i)]_{y_i} \le   \gamma + \max_{y'\neq y_i}[f_\matW(\vecx_i)]_{y'} + \frac{\epsilon}{2} \max_{k\in[K] , z=\pm1} \max_{\matP\succeq 0, \diag(\matP) \le 1} \innerps{zQ(\vecw_{2,k}, \matW_1)}{\matP}   \Big)  \\
& + \frac{1}{\gamma} \left( \frac{4}{n^{3/2}} + \frac{60\log(n)\log(2d_{\max})}{n}s_1s_2 \Big( (\frac{b_1}{s_1})^{2/3} + (\frac{b_2}{s_2})^{2/3} \Big)^{3/2} \fbnorms{\matX} + \frac{2\epsilon b_1b_2}{\sqrt{n}} \right) + 3\sqrt{\frac{\log \frac{2}{\delta}}{2n}}.
\end{align*}
\end{theorem}

We prove Theorem~\ref{thm:sdp_gen} in Appendix~\ref{prf:sdp_gen}. Similar to linear classifiers, in the adversarial setting, if we have an $\ell_1$ norm constraint on the matrix matrix $\matW_1$, then the generalization bound of the surrogate adversarial loss does not have an explicit dimension dependence.

\section{Experiments}\label{sec:experiments}
In this section, we validate our theoretical findings for linear classifiers and neural networks via experiments. Our experiments are implemented with Tensorflow~\cite{abadi2016tensorflow} on the MNIST dataset~\cite{lecun1998gradient}. \footnote{The implementation of the experiments can be found at \url{https://github.com/dongyin92/adversarially-robust-generalization}.}

\subsection{Linear Classifiers}
We validate two theoretical findings for linear classifiers: (i) controlling the $\ell_1$ norm of the model parameters can reduce the adversarial generalization error, and (ii) there is a dimension dependence in adversarial generalization, i.e., adversarially robust generalization is harder when the dimension of the feature space is higher. We train the multi-class linear classifier using the following objective function:

\begin{equation}\label{eq:objective}
\min_{\matW}\frac{1}{n}\sum_{i=1}^n \max_{\vecx_i' \in \Ball^\infty_{\vecx_i}(\epsilon)} \ell(f_\matW(\vecx_i'), y_i) + \lambda \onenms{\matW},
\end{equation}
where $\ell(\cdot)$ is cross entropy loss and $f_\matW (\vecx) \equiv \matW \vecx$. Since we focus on the generalization property, we use a small number of training data so that the generalization gap is more significant. More specifically, in each run of the training algorithm, we randomly sample $n=1000$ data points from the training set of MNIST as the training data, and run adversarial training to minimize the objective~\eqref{eq:objective}. Our training algorithm alternates between mini-batch stochastic gradient descent with respect to $\matW$ and computing adversarial examples on the chosen batch in each iteration. Here, we note that since we consider linear classifiers, the adversarial examples can be analytically computed according to Appendix~\ref{prf:multi_adv}.

\begin{figure}[h]
\centering 
\includegraphics[width=0.6\linewidth]{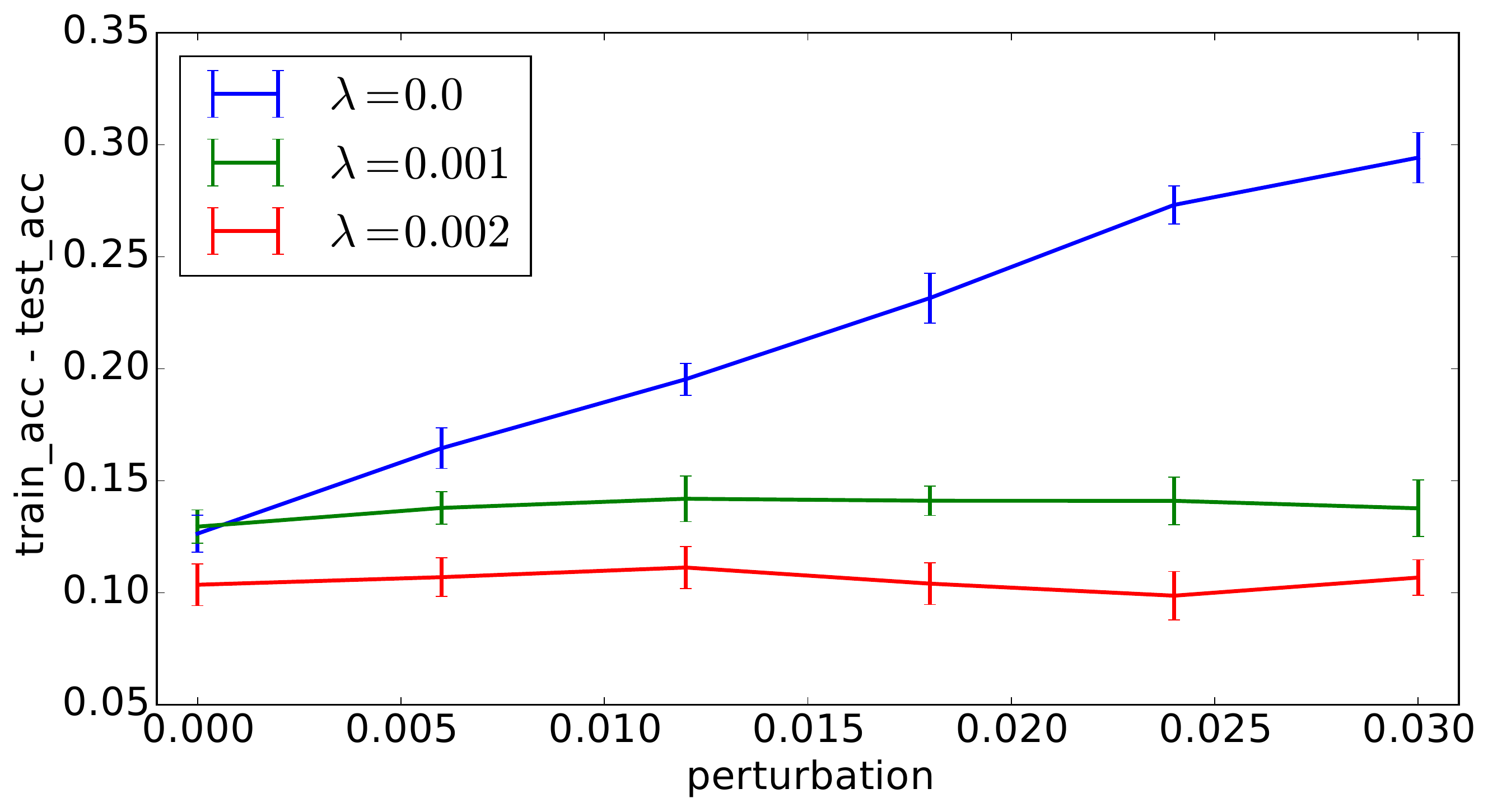}
\caption{Linear classifiers. Adversarial generalization error vs $\ell_\infty$ perturbation $\epsilon$ and regularization coefficient $\lambda$.}
\label{fig:gen_epsilon_lambda}
\end{figure}

In our first experiment, we vary the values of $\epsilon$ and $\lambda$, and for each $(\epsilon, \lambda)$ pair, we conduct $10$ runs of the training algorithm, and in each run we sample the $1000$ training data independently. In Figure~\ref{fig:gen_epsilon_lambda}, we plot the adversarial generalization error as a function of $\epsilon$ and $\lambda$, and the error bars show the standard deviation of the $10$ runs. As we can see, when $\lambda$ increases, the generalization gap decreases, and thus we conclude that $\ell_1$ regularization is helpful for reducing adversarial generalization error.

\begin{figure}[h]
\centering 
\includegraphics[width=0.6\linewidth]{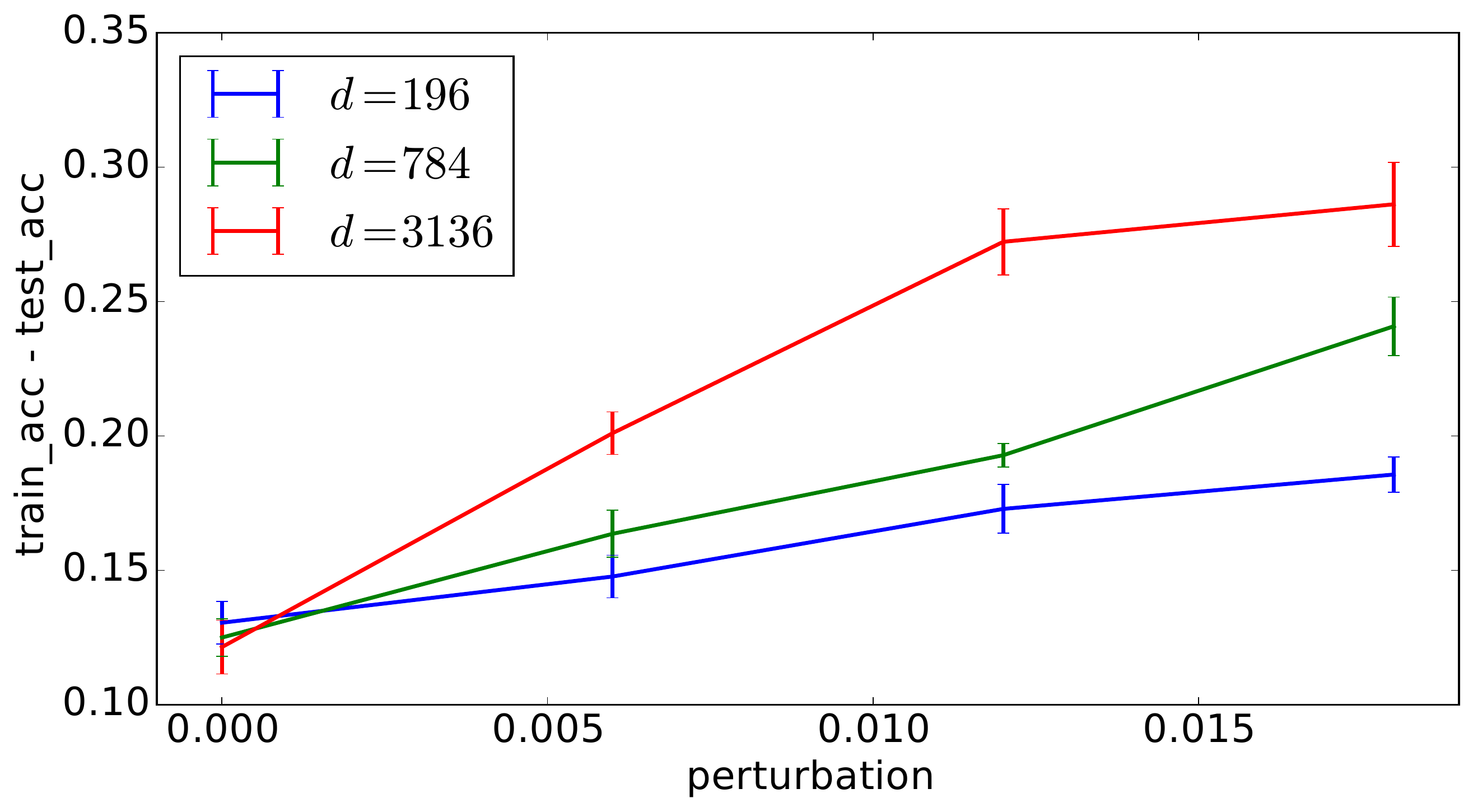}
\caption{Linear classifiers. Adversarial generalization error vs $\ell_\infty$ perturbation $\epsilon$ and dimension of feature space $d$.}
\label{fig:gen_epsilon_d}
\end{figure}

In our second experiment, we choose $\lambda=0$ and study the dependence of adversarial generalization error on the dimension of the feature space. Recall that each data point in the original MNIST dataset is a $28\times 28$ image, i.e., $d=784$. We construct two additional image datasets with $d=196$ (downsampled) and $d=3136$ (expanded), respectively. To construct the downsampled image, we replace each $2\times 2$ patch---say, with pixel values $a,b,c,d$---on the original image with a single pixel with value $\sqrt{a^2+b^2+c^2+d^2}$. To construct the expanded image, we replace each pixel---say, with value $a$---on the original image with a $2\times 2$ patch, with the value of each pixel in the patch being $a/2$. Such construction keeps the $\ell_2$ norm of the every single image the same across the three datasets, and thus leads a fair comparison. The adversarial generalization error is plotted in Figure~\ref{fig:gen_epsilon_d}, and as we can see, when the dimension $d$ increases, the generalization gap also increases.

\subsection{Neural Networks}
\begin{figure}[h]
\centering 
\includegraphics[width=0.6\linewidth]{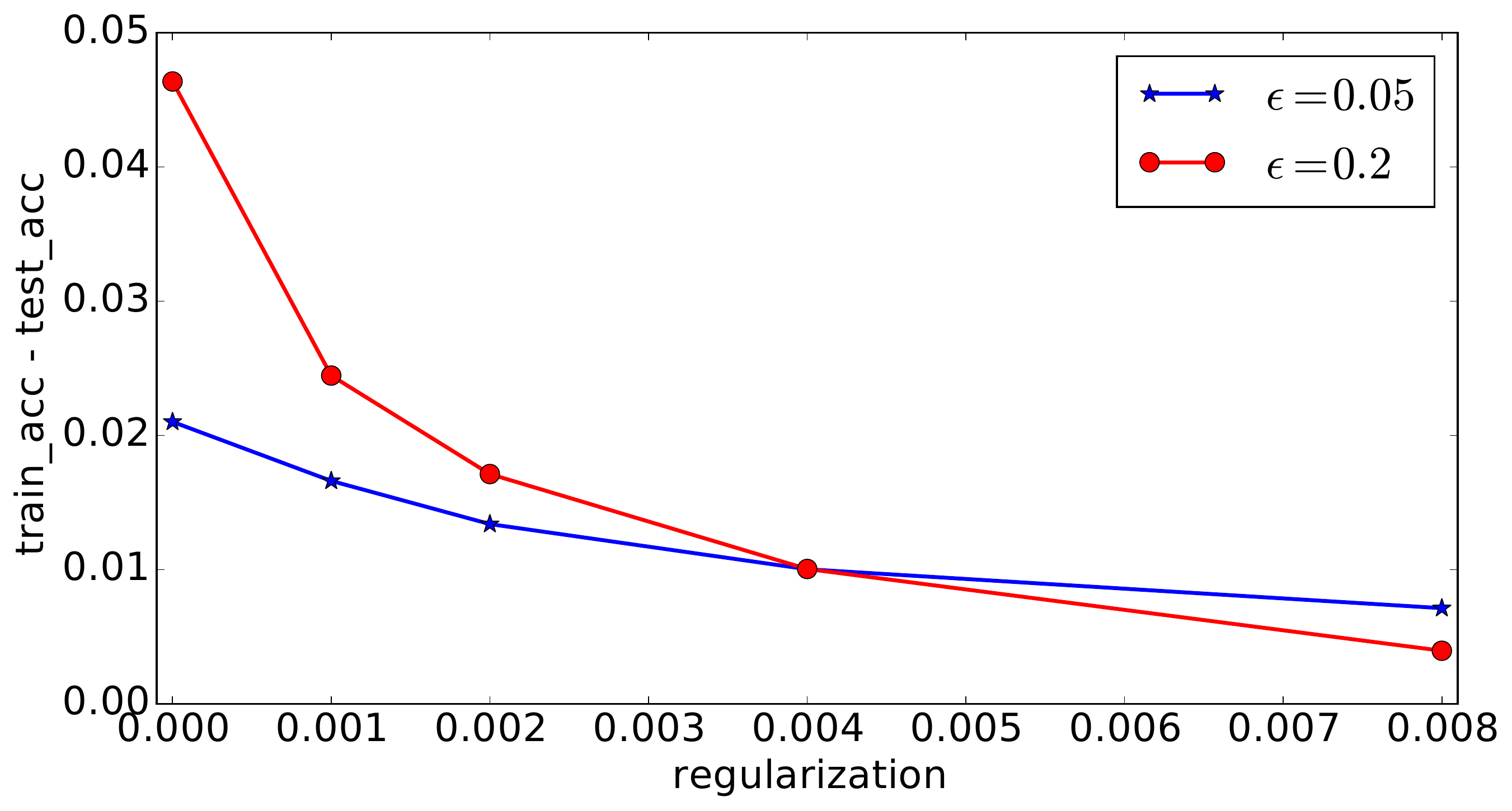}
\caption{Neural networks. Adversarial generalization error vs regularization coefficient $\lambda$.}
\label{fig:four_layer}
\end{figure}

In this experiment, we validate our theoretical result that $\ell_1$ regularization can reduce the adversarial generalization error on a four-layer ReLU neural network, where the first two layers are convolutional and the last two layers are fully connected. We use PGD attack~\cite{madry2017towards} adversarial training to minimize the $\ell_1$ regularized objective~\eqref{eq:objective}. We use the whole training set of MNIST, and once the model is obtained, we use PGD attack to check the adversarial training and test error. We present the adversarial generalization errors under the PGD attack in Figure~\ref{fig:four_layer}. As we can see, the adversarial generalization error decreases as we increase the regularization coefficient $\lambda$; thus $\ell_1$ regularization indeed reduces the adversarial generalization error under the PGD attack.

\section{Conclusions}\label{sec:conclusions}
We study the adversarially robust generalization properties of linear classifiers and neural networks through the lens of Rademacher complexity. For binary linear classifiers, we prove tight bounds for the adversarial Rademacher complexity, and show that in the adversarial setting, Rademacher complexity is never smaller than that in the natural setting, and it has an unavoidable dimension dependence, unless the weight vector has bounded $\ell_1$ norm. The results also extends to multi-class linear classifiers. For neural networks, we prove a lower bound of the Rademacher complexity of the adversarial loss function class and show that there is also an unavoidable dimension dependence due to $\ell_\infty$ adversarial attack. We further consider a surrogate adversarial loss and prove margin bound for this setting. Our results indicate that having $\ell_1$ norm constraints on the weight matrices might be a potential way to improve generalization in the adversarial setting. Our experimental results validate our theoretical findings.

\subsection*{Acknowledgements} 
D. Yin is partially supported by Berkeley DeepDrive Industry Consortium. K. Ramchandran is partially supported by NSF CIF award 1703678. P. Bartlett is partially supported by NSF grant IIS-1619362. The authors would like to thank Justin Gilmer for helpful discussion.

\bibliographystyle{plainnat}
\bibliography{ref}

\begin{thebibliography}{59}
\providecommand{\natexlab}[1]{#1}
\providecommand{\url}[1]{\texttt{#1}}
\expandafter\ifx\csname urlstyle\endcsname\relax
  \providecommand{\doi}[1]{doi: #1}\else
  \providecommand{\doi}{doi: \begingroup \urlstyle{rm}\Url}\fi

\bibitem[Abadi et~al.(2016)Abadi, Barham, Chen, Chen, Davis, Dean, Devin,
  Ghemawat, Irving, Isard, et~al.]{abadi2016tensorflow}
Mart{\'\i}n Abadi, Paul Barham, Jianmin Chen, Zhifeng Chen, Andy Davis, Jeffrey
  Dean, Matthieu Devin, Sanjay Ghemawat, Geoffrey Irving, Michael Isard, et~al.
\newblock Tensorflow: a system for large-scale machine learning.
\newblock In \emph{OSDI}, volume~16, pages 265--283, 2016.

\bibitem[Anthony and Bartlett(1999)]{anthony2009neural}
Martin Anthony and Peter~L Bartlett.
\newblock \emph{Neural network learning: Theoretical foundations}.
\newblock Cambridge University Press, 1999.

\bibitem[Arora et~al.(2018)Arora, Ge, Neyshabur, and Zhang]{arora2018stronger}
Sanjeev Arora, Rong Ge, Behnam Neyshabur, and Yi~Zhang.
\newblock Stronger generalization bounds for deep nets via a compression
  approach.
\newblock \emph{arXiv preprint arXiv:1802.05296}, 2018.

\bibitem[Athalye et~al.(2018)Athalye, Carlini, and
  Wagner]{athalye2018obfuscated}
Anish Athalye, Nicholas Carlini, and David Wagner.
\newblock Obfuscated gradients give a false sense of security: Circumventing
  defenses to adversarial examples.
\newblock \emph{arXiv preprint arXiv:1802.00420}, 2018.

\bibitem[Attias et~al.(2018)Attias, Kontorovich, and
  Mansour]{attias2018improved}
Idan Attias, Aryeh Kontorovich, and Yishay Mansour.
\newblock Improved generalization bounds for robust learning.
\newblock \emph{arXiv preprint arXiv:1810.02180}, 2018.

\bibitem[Bahdanau et~al.(2014)Bahdanau, Cho, and Bengio]{bahdanau2014neural}
Dzmitry Bahdanau, Kyunghyun Cho, and Yoshua Bengio.
\newblock Neural machine translation by jointly learning to align and
  translate.
\newblock \emph{arXiv preprint arXiv:1409.0473}, 2014.

\bibitem[Bartlett(1998)]{b-scpcnn-98}
Peter~L Bartlett.
\newblock The sample complexity of pattern classification with neural networks:
  the size of the weights is more important than the size of the network.
\newblock \emph{IEEE Transactions on Information Theory}, 44\penalty0
  (2):\penalty0 525--536, 1998.

\bibitem[Bartlett and Mendelson(2002)]{bartlett2002rademacher}
Peter~L Bartlett and Shahar Mendelson.
\newblock Rademacher and {G}aussian complexities: Risk bounds and structural
  results.
\newblock \emph{Journal of Machine Learning Research}, 3\penalty0
  (Nov):\penalty0 463--482, 2002.

\bibitem[Bartlett et~al.(2017)Bartlett, Foster, and
  Telgarsky]{bartlett2017spectrally}
Peter~L Bartlett, Dylan~J Foster, and Matus~J Telgarsky.
\newblock Spectrally-normalized margin bounds for neural networks.
\newblock In \emph{Advances in Neural Information Processing Systems}, 2017.

\bibitem[Ben-Tal et~al.(2009)Ben-Tal, El~Ghaoui, and Nemirovski]{ben2009robust}
Aharon Ben-Tal, Laurent El~Ghaoui, and Arkadi Nemirovski.
\newblock \emph{Robust optimization}.
\newblock Princeton University Press, 2009.

\bibitem[Bubeck et~al.(2018)Bubeck, Price, and
  Razenshteyn]{bubeck2018adversarial}
S{\'e}bastien Bubeck, Eric Price, and Ilya Razenshteyn.
\newblock Adversarial examples from computational constraints.
\newblock \emph{arXiv preprint arXiv:1805.10204}, 2018.

\bibitem[Carlini and Wagner(2016)]{carlini2016defensive}
Nicholas Carlini and David Wagner.
\newblock Defensive distillation is not robust to adversarial examples.
\newblock \emph{arXiv preprint arXiv:1607.04311}, 2016.

\bibitem[Carlini and Wagner(2017)]{carlini2017adversarial}
Nicholas Carlini and David Wagner.
\newblock Adversarial examples are not easily detected: Bypassing ten detection
  methods.
\newblock In \emph{Proceedings of the 10th ACM Workshop on Artificial
  Intelligence and Security}, pages 3--14. ACM, 2017.

\bibitem[Carlini and Wagner(2018)]{carlini2018audio}
Nicholas Carlini and David Wagner.
\newblock Audio adversarial examples: Targeted attacks on speech-to-text.
\newblock \emph{arXiv preprint arXiv:1801.01944}, 2018.

\bibitem[Cullina et~al.(2018)Cullina, Bhagoji, and Mittal]{cullina2018pac}
Daniel Cullina, Arjun~Nitin Bhagoji, and Prateek Mittal.
\newblock {PAC}-learning in the presence of evasion adversaries.
\newblock \emph{arXiv preprint arXiv:1806.01471}, 2018.

\bibitem[Dohmatob(2018)]{dohmatob2018limitations}
Elvis Dohmatob.
\newblock Limitations of adversarial robustness: strong no free lunch theorem.
\newblock \emph{arXiv preprint arXiv:1810.04065}, 2018.

\bibitem[Engstrom et~al.(2017)Engstrom, Tsipras, Schmidt, and
  Madry]{engstrom2017rotation}
Logan Engstrom, Dimitris Tsipras, Ludwig Schmidt, and Aleksander Madry.
\newblock A rotation and a translation suffice: Fooling {CNN}s with simple
  transformations.
\newblock \emph{arXiv preprint arXiv:1712.02779}, 2017.

\bibitem[Farnia et~al.(2018)Farnia, Zhang, and Tse]{farnia2018generalizable}
Farzan Farnia, Jesse~M Zhang, and David Tse.
\newblock Generalizable adversarial training via spectral normalization.
\newblock \emph{arXiv preprint arXiv:1811.07457}, 2018.

\bibitem[Fawzi et~al.(2016)Fawzi, Moosavi-Dezfooli, and
  Frossard]{fawzi2016robustness}
Alhussein Fawzi, Seyed-Mohsen Moosavi-Dezfooli, and Pascal Frossard.
\newblock Robustness of classifiers: from adversarial to random noise.
\newblock In \emph{Advances in Neural Information Processing Systems}, 2016.

\bibitem[Fawzi et~al.(2018)Fawzi, Fawzi, and Fawzi]{fawzi2018adversarial}
Alhussein Fawzi, Hamza Fawzi, and Omar Fawzi.
\newblock Adversarial vulnerability for any classifier.
\newblock \emph{arXiv preprint arXiv:1802.08686}, 2018.

\bibitem[Gilmer et~al.(2018{\natexlab{a}})Gilmer, Adams, Goodfellow, Andersen,
  and Dahl]{gilmer2018motivating}
Justin Gilmer, Ryan~P Adams, Ian Goodfellow, David Andersen, and George~E Dahl.
\newblock Motivating the rules of the game for adversarial example research.
\newblock \emph{arXiv preprint arXiv:1807.06732}, 2018{\natexlab{a}}.

\bibitem[Gilmer et~al.(2018{\natexlab{b}})Gilmer, Metz, Faghri, Schoenholz,
  Raghu, Wattenberg, and Goodfellow]{gilmer2018adversarial}
Justin Gilmer, Luke Metz, Fartash Faghri, Samuel~S Schoenholz, Maithra Raghu,
  Martin Wattenberg, and Ian Goodfellow.
\newblock Adversarial spheres.
\newblock \emph{arXiv preprint arXiv:1801.02774}, 2018{\natexlab{b}}.

\bibitem[Golowich et~al.(2017)Golowich, Rakhlin, and Shamir]{golowich2017size}
Noah Golowich, Alexander Rakhlin, and Ohad Shamir.
\newblock Size-independent sample complexity of neural networks.
\newblock \emph{arXiv preprint arXiv:1712.06541}, 2017.

\bibitem[Goodfellow et~al.(2014)Goodfellow, Shlens, and
  Szegedy]{goodfellow6572explaining}
Ian~J Goodfellow, Jonathon Shlens, and Christian Szegedy.
\newblock Explaining and harnessing adversarial examples.
\newblock \emph{arXiv preprint arXiv:1412.6572}, 2014.

\bibitem[Graves et~al.(2013)Graves, Mohamed, and Hinton]{graves2013speech}
Alex Graves, Abdel-rahman Mohamed, and Geoffrey Hinton.
\newblock Speech recognition with deep recurrent neural networks.
\newblock In \emph{ICASSP}. IEEE, 2013.

\bibitem[Gu and Rigazio(2014)]{gu2014towards}
Shixiang Gu and Luca Rigazio.
\newblock Towards deep neural network architectures robust to adversarial
  examples.
\newblock \emph{arXiv preprint arXiv:1412.5068}, 2014.

\bibitem[He et~al.(2016)He, Zhang, Ren, and Sun]{he2016deep}
Kaiming He, Xiangyu Zhang, Shaoqing Ren, and Jian Sun.
\newblock Deep residual learning for image recognition.
\newblock In \emph{Computer Vision and Pattern Recognition}, 2016.

\bibitem[Huang et~al.(2015)Huang, Xu, Schuurmans, and
  Szepesv{\'a}ri]{huang2015learning}
Ruitong Huang, Bing Xu, Dale Schuurmans, and Csaba Szepesv{\'a}ri.
\newblock Learning with a strong adversary.
\newblock \emph{arXiv preprint arXiv:1511.03034}, 2015.

\bibitem[Khim and Loh(2018)]{khim2018adversarial}
Justin Khim and Po-Ling Loh.
\newblock Adversarial risk bounds for binary classification via function
  transformation.
\newblock \emph{arXiv preprint arXiv:1810.09519}, 2018.

\bibitem[Koltchinskii et~al.(2006)]{koltchinskii2006local}
Vladimir Koltchinskii et~al.
\newblock Local rademacher complexities and oracle inequalities in risk
  minimization.
\newblock \emph{The Annals of Statistics}, 34\penalty0 (6):\penalty0
  2593--2656, 2006.

\bibitem[Kolter and Wong(2017)]{kolter2017provable}
J~Zico Kolter and Eric Wong.
\newblock Provable defenses against adversarial examples via the convex outer
  adversarial polytope.
\newblock \emph{arXiv preprint arXiv:1711.00851}, 2017.

\bibitem[Kos et~al.(2018)Kos, Fischer, and Song]{kos2018adversarial}
Jernej Kos, Ian Fischer, and Dawn Song.
\newblock Adversarial examples for generative models.
\newblock In \emph{2018 IEEE Security and Privacy Workshops (SPW)}, pages
  36--42. IEEE, 2018.

\bibitem[Kuznetsov et~al.(2015)Kuznetsov, Mohri, and
  Syed]{kuznetsov2015rademacher}
Vitaly Kuznetsov, Mehryar Mohri, and U~Syed.
\newblock Rademacher complexity margin bounds for learning with a large number
  of classes.
\newblock In \emph{ICML Workshop on Extreme Classification: Learning with a
  Very Large Number of Labels}, 2015.

\bibitem[LeCun et~al.(1998)LeCun, Bottou, Bengio, and
  Haffner]{lecun1998gradient}
Yann LeCun, L{\'e}on Bottou, Yoshua Bengio, and Patrick Haffner.
\newblock Gradient-based learning applied to document recognition.
\newblock \emph{Proceedings of the IEEE}, 86\penalty0 (11):\penalty0
  2278--2324, 1998.

\bibitem[Ledoux and Talagrand(2013)]{ledoux2013probability}
Michel Ledoux and Michel Talagrand.
\newblock \emph{Probability in {B}anach Spaces: isoperimetry and processes}.
\newblock Springer Science \& Business Media, 2013.

\bibitem[Lee et~al.(1996)Lee, Bartlett, and Williamson]{lee1996efficient}
Wee~Sun Lee, Peter~L Bartlett, and Robert~C Williamson.
\newblock Efficient agnostic learning of neural networks with bounded fan-in.
\newblock \emph{IEEE Transactions on Information Theory}, 42\penalty0
  (6):\penalty0 2118--2132, 1996.

\bibitem[Madry et~al.(2017)Madry, Makelov, Schmidt, Tsipras, and
  Vladu]{madry2017towards}
Aleksander Madry, Aleksandar Makelov, Ludwig Schmidt, Dimitris Tsipras, and
  Adrian Vladu.
\newblock Towards deep learning models resistant to adversarial attacks.
\newblock \emph{arXiv preprint arXiv:1706.06083}, 2017.

\bibitem[Mahloujifar et~al.(2018)Mahloujifar, Diochnos, and
  Mahmoody]{mahloujifar2018curse}
Saeed Mahloujifar, Dimitrios~I Diochnos, and Mohammad Mahmoody.
\newblock The curse of concentration in robust learning: Evasion and poisoning
  attacks from concentration of measure.
\newblock \emph{arXiv preprint arXiv:1809.03063}, 2018.

\bibitem[Maximov and Reshetova(2016)]{maximov2016tight}
Yu~Maximov and Daria Reshetova.
\newblock Tight risk bounds for multi-class margin classifiers.
\newblock \emph{Pattern Recognition and Image Analysis}, 26\penalty0
  (4):\penalty0 673--680, 2016.

\bibitem[Mei et~al.(2018)Mei, Montanari, and Nguyen]{mei2018mean}
Song Mei, Andrea Montanari, and Phan-Minh Nguyen.
\newblock A mean field view of the landscape of two-layers neural networks.
\newblock \emph{arXiv preprint arXiv:1804.06561}, 2018.

\bibitem[Mohri et~al.(2012)Mohri, Rostamizadeh, and
  Talwalkar]{mohri2012foundations}
Mehryar Mohri, Afshin Rostamizadeh, and Ameet Talwalkar.
\newblock \emph{Foundations of machine learning}.
\newblock MIT press, 2012.

\bibitem[Neyshabur et~al.(2017)Neyshabur, Bhojanapalli, McAllester, and
  Srebro]{neyshabur2017pac}
Behnam Neyshabur, Srinadh Bhojanapalli, David McAllester, and Nathan Srebro.
\newblock A {PAC}-bayesian approach to spectrally-normalized margin bounds for
  neural networks.
\newblock \emph{arXiv preprint arXiv:1707.09564}, 2017.

\bibitem[Papernot et~al.(2016)Papernot, McDaniel, Sinha, and
  Wellman]{papernot2016towards}
Nicolas Papernot, Patrick McDaniel, Arunesh Sinha, and Michael Wellman.
\newblock Towards the science of security and privacy in machine learning.
\newblock \emph{arXiv preprint arXiv:1611.03814}, 2016.

\bibitem[Raghunathan et~al.(2018{\natexlab{a}})Raghunathan, Steinhardt, and
  Liang]{raghunathan2018certified}
Aditi Raghunathan, Jacob Steinhardt, and Percy Liang.
\newblock Certified defenses against adversarial examples.
\newblock \emph{arXiv preprint arXiv:1801.09344}, 2018{\natexlab{a}}.

\bibitem[Raghunathan et~al.(2018{\natexlab{b}})Raghunathan, Steinhardt, and
  Liang]{raghunathan2018semidefinite}
Aditi Raghunathan, Jacob Steinhardt, and Percy~S Liang.
\newblock Semidefinite relaxations for certifying robustness to adversarial
  examples.
\newblock In \emph{Advances in Neural Information Processing Systems},
  2018{\natexlab{b}}.

\bibitem[Schmidt et~al.(2018)Schmidt, Santurkar, Tsipras, Talwar, and
  Madry]{schmidt2018adversarially}
Ludwig Schmidt, Shibani Santurkar, Dimitris Tsipras, Kunal Talwar, and
  Aleksander Madry.
\newblock Adversarially robust generalization requires more data.
\newblock \emph{arXiv preprint arXiv:1804.11285}, 2018.

\bibitem[Shaham et~al.(2015)Shaham, Yamada, and
  Negahban]{shaham2015understanding}
Uri Shaham, Yutaro Yamada, and Sahand Negahban.
\newblock Understanding adversarial training: Increasing local stability of
  neural nets through robust optimization.
\newblock \emph{arXiv preprint arXiv:1511.05432}, 2015.

\bibitem[Silver et~al.(2016)Silver, Huang, Maddison, Guez, Sifre, Van
  Den~Driessche, Schrittwieser, Antonoglou, Panneershelvam, Lanctot,
  et~al.]{silver2016mastering}
David Silver, Aja Huang, Chris~J Maddison, Arthur Guez, Laurent Sifre, George
  Van Den~Driessche, Julian Schrittwieser, Ioannis Antonoglou, Veda
  Panneershelvam, Marc Lanctot, et~al.
\newblock Mastering the game of go with deep neural networks and tree search.
\newblock \emph{Nature}, 529\penalty0 (7587):\penalty0 484, 2016.

\bibitem[Sinha et~al.(2018)Sinha, Namkoong, and Duchi]{sinha2018certifying}
Aman Sinha, Hongseok Namkoong, and John Duchi.
\newblock Certifying some distributional robustness with principled adversarial
  training.
\newblock In \emph{International Conference on Learning Representations}, 2018.

\bibitem[Suggala et~al.(2018)Suggala, Prasad, Nagarajan, and
  Ravikumar]{suggala2018adversarial}
Arun~Sai Suggala, Adarsh Prasad, Vaishnavh Nagarajan, and Pradeep Ravikumar.
\newblock On adversarial risk and training.
\newblock \emph{arXiv preprint arXiv:1806.02924}, 2018.

\bibitem[Szegedy et~al.(2013)Szegedy, Zaremba, Sutskever, Bruna, Erhan,
  Goodfellow, and Fergus]{szegedy2013intriguing}
Christian Szegedy, Wojciech Zaremba, Ilya Sutskever, Joan Bruna, Dumitru Erhan,
  Ian Goodfellow, and Rob Fergus.
\newblock Intriguing properties of neural networks.
\newblock \emph{arXiv preprint arXiv:1312.6199}, 2013.

\bibitem[Tsipras et~al.(2018)Tsipras, Santurkar, Engstrom, Turner, and
  Madry]{tsipras2018there}
Dimitris Tsipras, Shibani Santurkar, Logan Engstrom, Alexander Turner, and
  Aleksander Madry.
\newblock Robustness may be at odds with accuracy.
\newblock \emph{arXiv preprint arXiv:1805.12152}, 2018.

\bibitem[Wang et~al.(2017)Wang, Jha, and Chaudhuri]{wang2017analyzing}
Yizhen Wang, Somesh Jha, and Kamalika Chaudhuri.
\newblock Analyzing the robustness of nearest neighbors to adversarial
  examples.
\newblock \emph{arXiv preprint arXiv:1706.03922}, 2017.

\bibitem[Wong et~al.(2018)Wong, Schmidt, Metzen, and Kolter]{wong2018scaling}
Eric Wong, Frank Schmidt, Jan~Hendrik Metzen, and J~Zico Kolter.
\newblock Scaling provable adversarial defenses.
\newblock \emph{arXiv preprint arXiv:1805.12514}, 2018.

\bibitem[Xu and Mannor(2012)]{xu2012robustness}
Huan Xu and Shie Mannor.
\newblock Robustness and generalization.
\newblock \emph{Machine learning}, 86\penalty0 (3):\penalty0 391--423, 2012.

\bibitem[Xu et~al.(2009{\natexlab{a}})Xu, Caramanis, and Mannor]{xu2009robust}
Huan Xu, Constantine Caramanis, and Shie Mannor.
\newblock Robust regression and {L}asso.
\newblock In \emph{Advances in Neural Information Processing Systems},
  2009{\natexlab{a}}.

\bibitem[Xu et~al.(2009{\natexlab{b}})Xu, Caramanis, and
  Mannor]{xu2009robustness}
Huan Xu, Constantine Caramanis, and Shie Mannor.
\newblock Robustness and regularization of support vector machines.
\newblock \emph{JMLR}, 10\penalty0 (Jul):\penalty0 1485--1510,
  2009{\natexlab{b}}.

\bibitem[Zhang et~al.(2016{\natexlab{a}})Zhang, Bengio, Hardt, Recht, and
  Vinyals]{zhang2016understanding}
Chiyuan Zhang, Samy Bengio, Moritz Hardt, Benjamin Recht, and Oriol Vinyals.
\newblock Understanding deep learning requires rethinking generalization.
\newblock \emph{arXiv preprint arXiv:1611.03530}, 2016{\natexlab{a}}.

\bibitem[Zhang et~al.(2016{\natexlab{b}})Zhang, Lee, and Jordan]{zhang2016l1}
Yuchen Zhang, Jason~D Lee, and Michael~I Jordan.
\newblock $\ell_1$-regularized neural networks are improperly learnable in
  polynomial time.
\newblock In \emph{International Conference on Machine Learning},
  2016{\natexlab{b}}.

\end{thebibliography}

\appendix

\section{Proof of Theorem~\ref{thm:rad_compare}}\label{prf:rad_compare}
First, we have
\begin{equation}\label{eq:rad_vanilla}
\frakR_{\setS}(\F) := \frac{1}{n}\EE_{\vecsigma} \left[ \sup_{\pnms{\vecw} \le W} \sum_{i=1}^n \sigma_i \innerps{\vecw}{\vecx_i} \right] = \frac{W}{n} \EE_{\vecsigma} \left[ \qnm{\sum_{i=1}^n \sigma_i \vecx_i} \right].
\end{equation}
We then analyze $\frakR_{\setS}(\tcalF)$. Define $\tf_\vecw(\vecx, y):= \min_{\vecx' \in \Ball^\infty_\vecx(\epsilon)} y\innerps{\vecw}{\vecx'}$. Then, we have
\[
\tf_\vecw(\vecx, y) = \begin{cases}
 \min_{\vecx' \in \Ball^\infty_\vecx(\epsilon)} \innerps{\vecw}{\vecx'} & y=1, \\
 - \max_{\vecx' \in \Ball^\infty_\vecx(\epsilon)} \innerps{\vecw}{\vecx'} & y = -1.
\end{cases}
\]
When $y=1$, we have
\begin{align*}
\tf_\vecw(\vecx, y) &= \tf_\vecw(\vecx, 1) = \min_{ \vecx' \in \Ball^\infty_\vecx(\epsilon) } \innerps{\vecw}{\vecx'} = \min_{ \vecx' \in \Ball^\infty_\vecx(\epsilon) } \sum_{i=1}^d w_i x'_i  \\
&= \sum_{i=1}^d w_i \left[  \indi(w_i \ge 0) (x_i - \epsilon) + \indi(w_i < 0)(x_i + \epsilon)  \right] = \sum_{i=1}^d w_i (x_i - \sgn(w_i)\epsilon)  \\
&= \innerps{\vecw}{\vecx} - \epsilon \onenms{\vecw}.
\end{align*}
Similarly, when $y=-1$, we have
\begin{align*}
\tf_\vecw(\vecx, y) &= \tf_\vecw(\vecx, -1) = -\max_{ \vecx' \in \Ball^\infty_\vecx(\epsilon) } \innerps{\vecw}{\vecx'} = - \max_{ \vecx' \in \Ball^\infty_\vecx(\epsilon) } \sum_{i=1}^d w_i x'_i  \\
&= - \sum_{i=1}^d w_i \left[  \indi(w_i \ge 0) (x_i + \epsilon) + \indi(w_i < 0)(x_i - \epsilon)  \right] = -\sum_{i=1}^d w_i (x_i + \sgn(w_i)\epsilon)  \\
&= -\innerps{\vecw}{\vecx} - \epsilon \onenms{\vecw}.
\end{align*}
Thus, we conclude that $\tf_\vecw(\vecx, y) = y\innerps{\vecw}{\vecx} - \epsilon \onenms{\vecw}$, and therefore
\[
\frakR_{\setS}(\tcalF) = \frac{1}{n} \EE_{\vecsigma} \left[ \sup_{\twonms{\vecw} \le W} \sum_{i=1}^n \sigma_i ( y_i\innerps{\vecw}{\vecx_i}  - \epsilon \onenms{\vecw})\right].
\]
Define $\vecu := \sum_{i=1}^n \sigma_i y_i \vecx_i$ and $v := \epsilon\sum_{i=1}^n \sigma_i$. Then we have
\[
\frakR_{\setS}(\tcalF) = \frac{1}{n} \EE_{\vecsigma} \left[ \sup_{\pnms{\vecw} \le W} \innerps{\vecw}{\vecu} - v\onenms{\vecw} \right]
\]
Since the supremum of $ \innerps{\vecw}{\vecu} - v\onenms{\vecw}$ over $\vecw$ can only be achieved when $\sgn(w_i) = \sgn(u_i)$, we know that
\begin{align}
\frakR_{\setS}(\tcalF) = & \frac{1}{n} \EE_{\vecsigma} \left[ \sup_{\pnms{\vecw} \le W} \innerps{\vecw}{\vecu} - v\innerps{\vecw}{\sgn(\vecw)} \right] \nonumber \\
 = & \frac{1}{n} \EE_{\vecsigma} \left[ \sup_{\pnms{\vecw} \le W} \innerps{\vecw}{\vecu} - v\innerps{\vecw}{\sgn(\vecu)} \right]  \nonumber  \\
 = & \frac{1}{n} \EE_{\vecsigma} \left[  \sup_{\pnms{\vecw} \le W}  \innerps{\vecw}{\vecu - v \sgn(\vecu)}  \right]  \nonumber  \\
 = & \frac{W}{n} \EE_{\vecsigma} \left[ \qnms{\vecu - v \sgn(\vecu)} \right]   \nonumber  \\
 = & \frac{W}{n} \EE_{\vecsigma} \left[  \qnm{ \sum_{i=1}^n \sigma_i y_i \vecx_i  - \big( \epsilon\sum_{i=1}^n \sigma_i \big) \sgn( \sum_{i=1}^n \sigma_i y_i \vecx_i )   }  \right]. \label{eq:rad_norm}
\end{align}
Now we prove an upper bound for $\frakR_{\setS}(\tcalF)$. By triangle inequality, we have
\begin{align*}
\frakR_{\setS}(\tcalF) \le & \frac{W}{n} \EE_{\vecsigma} \left[  \qnm{ \sum_{i=1}^n \sigma_i y_i \vecx_i } \right] + \frac{\epsilon W}{n} \EE_{\vecsigma} \left[ \qnm{ (\sum_{i=1}^n \sigma_i ) \sgn( \sum_{i=1}^n \sigma_i y_i \vecx_i ) } \right] \\
= & \frakR_{\setS}(\F) + \epsilon W \frac{ d^{\frac{1}{q}} }{n} \EE_{\vecsigma} \left[ \abs{\sum_{i=1}^n \sigma_i}  \right] \\
\le & \frakR_{\setS}(\F) + \epsilon W\frac{d^{\frac{1}{q}}}{\sqrt{n}},
\end{align*}
where the last step is due to Khintchine's inequality.

We then proceed to prove a lower bound for $\frakR_{\setS}(\tcalF)$. According to~\eqref{eq:rad_norm} and by symmetry, we know that
\begin{align}
\frakR_{\setS}(\tcalF) = & \frac{W}{n} \EE_{\vecsigma} \left[  \qnm{ \sum_{i=1}^n (-\sigma_i) y_i \vecx_i  -  \big( \epsilon\sum_{i=1}^n (-\sigma_i) \big) \sgn( \sum_{i=1}^n (-\sigma_i) y_i \vecx_i )   }  \right]  \nonumber  \\
= & \frac{W}{n} \EE_{\vecsigma} \left[  \qnm{ \sum_{i=1}^n \sigma_i y_i \vecx_i  +  \big( \epsilon\sum_{i=1}^n \sigma_i \big) \sgn( \sum_{i=1}^n \sigma_i y_i \vecx_i )   }  \right].  \label{eq:rad_norm2}
\end{align}
Then, combining~\eqref{eq:rad_norm} and~\eqref{eq:rad_norm2} and using triangle inequality, we have
\begin{align}
\frakR_{\setS}(\tcalF) = & \frac{W}{2n} \EE_{\vecsigma} \Bigg[  \qnm{ \sum_{i=1}^n \sigma_i y_i \vecx_i  - \big( \epsilon\sum_{i=1}^n \sigma_i \big) \sgn( \sum_{i=1}^n \sigma_i y_i \vecx_i )   } \nonumber \\
& +  \qnm{ \sum_{i=1}^n \sigma_i y_i \vecx_i  +  \big( \epsilon\sum_{i=1}^n \sigma_i \big) \sgn( \sum_{i=1}^n \sigma_i y_i \vecx_i )   } \Bigg]  \nonumber \\
\ge & \frac{W}{n}  \EE_{\vecsigma} \left[ \qnm{ \sum_{i=1}^n \sigma_i y_i \vecx_i } \right] = \frakR_{\setS}(\F).  \label{eq:rad_adv_lower1}
\end{align}
Similarly, we have
\begin{align*}
\frakR_{\setS}(\tcalF) \ge & \frac{W}{n}  \EE_{\vecsigma} \left[ \qnm{ \big( \epsilon\sum_{i=1}^n \sigma_i \big) \sgn( \sum_{i=1}^n \sigma_i y_i \vecx_i )   } \right]  \\
= & \frac{W}{n}  \EE_{\vecsigma} \left[ \epsilon \left| \sum_{i=1}^n \sigma_i \right| \qnm{\sgn( \sum_{i=1}^n \sigma_i y_i \vecx_i )}  \right]  \\
= & \epsilon W \frac{d^{\frac{1}{q}}}{n }\EE_{\vecsigma} \left[ \abs{ \sum_{i=1}^n \sigma_i } \right].
\end{align*}
By Khintchine's inequality, we know that there exists a universal constant $c>0$ such that $\EE_{\vecsigma} [ \abs{ \sum_{i=1}^n \sigma_i } ] \ge c\sqrt{n}$. Therefore, we have $ \frakR_{\setS}(\tcalF) \ge c\epsilon W \frac{d^{\frac{1}{q}}}{\sqrt{n}} $. Combining with~\eqref{eq:rad_adv_lower1}, we complete the proof.

\section{Multi-class Linear Classifiers}

\subsection{Proof of Theorem~\ref{thm:multi_nat}}\label{prf:multi_nat}
According to the multi-class margin bound in~\cite{kuznetsov2015rademacher}, for any fixed $\gamma$, with probability at least $1-\delta$, we have
\begin{align*}
\PP_{(\vecx, y)\sim\D} \left\{ y \neq \arg\max_{y' \in [K]} [f(\vecx)]_{y'} \right\} \le & \frac{1}{n}\sum_{i=1}^n \indi([f(\vecx_i)]_{y_i} \le \gamma + \max_{y' \neq y}[f(\vecx_i)]_{y'}) \\
& + \frac{4K}{\gamma}\frakR_{\setS}(\Pi_1(\F)) + 3\sqrt{\frac{\log \frac{2}{\delta}}{2n}},
\end{align*}
where $\Pi_1(\F) \subseteq \R^{\X}$ is defined as
\[
\Pi_1(\F) := \{\vecx \mapsto [f(\vecx)]_k : f\in\F, k\in[K] \}.
\]
In the special case of linear classifiers $\F = \{ f_\matW(\vecx) : \norms{\matW^\top}_{p,\infty} \le W \}$, we can see that
\[
\Pi_1(\F) = \{\vecx \mapsto \innerps{\vecw}{\vecx} : \pnms{\vecw} \le W\}.
\]
Thus, we have
\[
\frakR_{\setS}(\Pi_1(\F)) = \frac{1}{n} \EE_{\vecsigma}\left[ \qnm{\sum_{i=1}^n \sigma_i\vecx_i } \right],
\]
which completes the proof.

\subsection{Proof of Theorem~\ref{thm:multi_adv}}\label{prf:multi_adv}
Since the loss function in the adversarial setting is 
\[
\tL(f_{\matW}(\vecx), y) = \max_{ \vecx' \in \Ball^\infty_\vecx(\epsilon) } \phi_\gamma( M(f_{\matW}(\vecx), y) ) = \phi_\gamma(\min_{ \vecx' \in \Ball^\infty_\vecx(\epsilon) } M(f_{\matW}(\vecx), y) ).
\]
Since we consider linear classifiers, we have
\begin{align}
\min_{ \vecx' \in \Ball^\infty_\vecx(\epsilon) } M(f_{\matW}(\vecx), y) &= \min_{ \vecx' \in \Ball^\infty_\vecx(\epsilon) } \min_{y' \neq y} (\vecw_y - \vecw_{y'})^\top \vecx'  \nonumber \\
&= \min_{y' \neq y} \min_{ \vecx' \in \Ball^\infty_\vecx(\epsilon) } (\vecw_y - \vecw_{y'})^\top \vecx'  \nonumber \\
&= \min_{y' \neq y} (\vecw_y - \vecw_{y'})^\top \vecx - \epsilon \onenms{\vecw_y - \vecw_{y'}}  \label{eq:min_margin}
\end{align}
Define
\[
h^{(k)}_{\matW}(\vecx, y) := (\vecw_y - \vecw_k)^\top \vecx - \epsilon \onenms{\vecw_y - \vecw_k} + \gamma \indi(y=k).
\]
We now show that 
\begin{equation}\label{eq:max_hw}
\tL(f_{\matW}(\vecx), y) = \max_{k\in[K]} \phi_\gamma (h^{(k)}_{\matW}(\vecx, y)).
\end{equation} 
To see this, we can see that according to~\eqref{eq:min_margin},
\[
\min_{\vecx' \in \Ball^\infty_\vecx(\epsilon)} M(f_{\matW}(\vecx), y) = \min_{k\neq y} h^{(k)}_{\matW}(\vecx, y).
\]
If $\min_{k\neq y} h^{(k)}_{\matW}(\vecx, y) \le \gamma$, we have $\min_{k\neq y} h^{(k)}_{\matW}(\vecx, y) = \min_{k\in [K]} h^{(k)}_{\matW}(\vecx, y)$, since $h^{(y)}_{\matW}(\vecx, y) = \gamma$. On the other hand, if $\min_{k\neq y} h^{(k)}_{\matW}(\vecx, y) > \gamma$, then $\min_{k\in [K]} h^{(k)}_{\matW}(\vecx, y) = \gamma$. In this case, we have $\phi_\gamma(\min_{k\neq y} h^{(k)}_{\matW}(\vecx, y)) = \phi_\gamma(\min_{k\in [K]} h^{(k)}_{\matW}(\vecx, y)) = 0$. Therefore, we can see that~\eqref{eq:max_hw} holds.

Define the $K$ function classes $\F_k := \{ h^{(k)}_{\matW}(\vecx, y) : \| \matW^\top \|_{p, \infty} \le W \} \subseteq \R^{\X \times \Y}$. Since $\phi_\gamma(\cdot)$ is $1/\gamma$-Lipschitz, according to the Ledoux-Talagrand contraction inequality~\cite{ledoux2013probability} and Lemma~8.1 in~\cite{mohri2012foundations}, we have
\begin{equation}\label{eq:max_sum}
\frakR_\setS (\tL_\F) \le \frac{1}{\gamma} \sum_{k=1}^K \frakR_\setS (\F_k).
\end{equation}
We proceed to analyze $\frakR_\setS(\F_k)$. The basic idea is similar to the proof of Theorem~\ref{thm:rad_compare}. We define $\vecu_y = \sum_{i=1}^n \sigma_i \vecx_i \indi( y_i = y )$ and $v_y = \sum_{i=1}^n \sigma_i \indi( y_i = y )$. Then, we have
\begin{align*}
\frakR_\setS (\F_k) = & \frac{1}{n} \EE_{\vecsigma} \Big[ \sup_{\| \matW^\top \|_{p, \infty} \le W}  \sum_{i=1}^n \sigma_i ( (\vecw_{y_i} - \vecw_k)^\top \vecx_i - \epsilon \onenms{\vecw_{y_i} - \vecw_k} + \gamma\indi(y_i=k) ) \Big]  \\
=& \frac{1}{n} \EE_{\vecsigma}  \Big[ \sup_{\| \matW^\top \|_{p, \infty} \le W}  \sum_{i=1}^n \sum_{y=1}^K \sigma_i ( (\vecw_{y_i} - \vecw_k)^\top \vecx_i - \epsilon \onenms{\vecw_{y_i} - \vecw_k} + \gamma\indi(y_i=k) ) \indi(y_i=y) \Big]   \\
=& \frac{1}{n} \EE_{\vecsigma}  \Big[ \sup_{\| \matW^\top \|_{p, \infty} \le W}  \sum_{y=1}^K  \sum_{i=1}^n  \sigma_i ( (\vecw_{y} - \vecw_k)^\top \vecx_i \indi(y_i=y) - \epsilon \onenms{\vecw_{y} - \vecw_k} \indi(y_i=y) \\
& + \gamma\indi(y_i=k) \indi(y_i=y) )  \Big]   \\
=& \frac{1}{n} \EE_{\vecsigma}  \Big[ \gamma\sum_{i=1}^n \sigma_i\indi(y_i = k) + \sup_{\| \matW^\top \|_{p, \infty} \le W}  \sum_{y\neq k} (\innerps{\vecw_y - \vecw_k}{\vecu_y} - \epsilon v_y \onenms{\vecw_y - \vecw_k}) \Big]   \\
\le & \frac{1}{n} \EE_{\vecsigma} \Big[ \sum_{y\neq k} \sup_{\pnms{\vecw_k},\pnms{\vecw_y} \le W} (\innerps{\vecw_y - \vecw_k}{\vecu_y} - \epsilon v_y \onenms{\vecw_y - \vecw_k}) \Big]  \\
= & \frac{1}{n} \EE_{\vecsigma} \Big[ \sum_{y\neq k} \sup_{\pnms{\vecw} \le 2W} (\innerps{\vecw}{\vecu_y} - \epsilon v_y \onenms{\vecw}) \Big] \\
= & \frac{2W}{n}  \EE_{\vecsigma} \Big[  \sum_{y\neq k} \qnms{\vecu_y - \epsilon v_y \sgn(\vecu_y)} \Big],
\end{align*}
where the last equality is due to the same derivation as in the proof of Theorem~\ref{thm:rad_compare}. Let $ n_y = \sum_{i=1}^n \indi(y_i = y) $. Then, we apply triangle inequality and Khintchine's inequality and obtain
\[
\frakR_\setS (\F_k) \le  \frac{2W}{n} \sum_{y\neq k} \EE_{\vecsigma} [ \twonms{\vecu_y} ] + \epsilon d^{\frac{1}{q}} \sqrt{n_y}.
\]
Combining with~\eqref{eq:max_sum}, we obtain
\begin{align*}
\frakR_\setS (\tL_\F) \le \frac{2WK}{\gamma n} (\sum_{y=1}^K \EE_{\vecsigma} [ \twonms{\vecu_y} ] + \epsilon d^{\frac{1}{q}} \sqrt{n_y} ) \le \frac{2WK}{\gamma} \left[ \frac{\epsilon  \sqrt{K}  d^{\frac{1}{q}} }{\sqrt{n}} + \frac{1}{n} \sum_{y=1}^K \EE_{\vecsigma} [ \twonms{\vecu_y}]  \right],
\end{align*}
where the last step is due to Cauchy-Schwarz inequality.

\section{Neural Network}
\subsection{Proof of Theorem~\ref{thm:nn_adv}}\label{prf:nn_adv}
We first review a Rademacher complexity lower bound in~\cite{bartlett2017spectrally}.
\begin{lemma}\label{lem:rad_nn_lower_nat}
~\cite{bartlett2017spectrally} Define the function class
\[
\widehat{\F} = \{ \vecx \mapsto f_\matW(\vecx) : \matW = (\matW_1,\matW_2,\ldots, \matW_L), \prod_{h=1}^L \| \matW_h \|_\sigma \le r  \},
\]
and $\widehat{\F}' = \{ \vecx \mapsto \innerps{\vecw}{\vecx} : \twonms{\vecw} \le \frac{r}{2} \}$. Then we have $ \widehat{\F}' \subseteq \widehat{\F}$, and thus there exists a universal constant $c>0$ such that
\[
\frakR_\setS (\widehat{\F}) \ge \frac{c r}{n} \fbnorms{\matX}.
\]
\end{lemma}
According to Lemma~\ref{lem:rad_nn_lower_nat}, in the adversarial setting, by defining
\[
\tcalF' = \{ \vecx \mapsto \min_{ \vecx' \in \Ball^\infty_\vecx(\epsilon) } y\innerps{\vecw}{\vecx'} : \twonms{\vecw} \le \frac{r}{2} \} \subseteq \R^{\X \times \{-1, +1\}},
\]
we have $\tcalF' \subseteq \tcalF$. Therefore, there exists a universal constant $c>0$ such that
\[
\frakR_\setS (\tcalF ) \ge \frakR_\setS( \tcalF' ) \ge cr \left(\frac{1}{n} \fbnorms{\matX} + \epsilon\sqrt{\frac{d}{n}} \right),
\]
where the last inequality is due to Theorem~\ref{thm:rad_compare}.

\subsection{Proof of Lemma~\ref{lem:surrogate}}\label{prf:surrogate}
Since $Q(\cdot, \cdot)$ is a linear function in its first argument, we have for any $y, y' \in [K]$,
\begin{align}
 & \max_{\matP\succeq 0, \diag(\matP) \le 1} \innerps{Q(\vecw_{2,y'} - \vecw_{2,y}, \matW_1)}{\matP}   \nonumber \\
\le &  \max_{\matP\succeq 0, \diag(\matP) \le 1} \innerps{Q(\vecw_{2,y'}, \matW_1)}{\matP} + \max_{\matP\succeq 0, \diag(\matP) \le 1} \innerps{-Q(\vecw_{2,y}, \matW_1)}{\matP}  \nonumber \\
\le & 2\max_{k\in[K] , z=\pm1} \max_{\matP\succeq 0, \diag(\matP) \le 1}\innerps{zQ(\vecw_{2,k}, \matW_1)}{\matP}.  \label{eq:max_k_z}
\end{align}
Then, for any $(\vecx, y)$, we have
\begin{align*}
&\max_{\vecx' \in \Ball^\infty_\vecx(\epsilon)} \indi(y \neq \arg\max_{y'\in[K]} [f_\matW(\vecx')]_{y'} ) \\
\le & \phi_\gamma(\min_{\vecx' \in \Ball^\infty_\vecx(\epsilon)} M(f_\matW(\vecx') , y))  \\
\le & \phi_\gamma(\min_{y' \neq y} \min_{\vecx' \in \Ball^\infty_\vecx(\epsilon)} [f_\matW(\vecx')]_{y} - [f_\matW(\vecx')]_{y'} ) \\
\le & \phi_\gamma \Big(\min_{y' \neq y} [f_\matW(\vecx)]_{y} - [f_\matW(\vecx)]_{y'} -  \frac{\epsilon}{4}\max_{y' \neq y}\max_{\matP\succeq 0, \diag(\matP) \le 1} \innerps{Q(\vecw_{2,y'} - \vecw_{2,y}, \matW_1)}{\matP} \Big) \\
\le & \phi_\gamma \Big( \min_{y' \neq y} [f_\matW(\vecx)]_{y} - [f_\matW(\vecx)]_{y'} - \frac{\epsilon}{2} \max_{k\in[K] , z=\pm1} \max_{\matP\succeq 0, \diag(\matP) \le 1} \innerps{zQ(\vecw_{2,k}, \matW_1)}{\matP}  \Big)  \\
\le & \phi_\gamma \Big( M(f_\matW(\vecx) , y) - \frac{\epsilon}{2} \max_{k\in[K] , z=\pm1} \max_{\matP\succeq 0, \diag(\matP) \le 1} \innerps{zQ(\vecw_{2,k}, \matW_1)}{\matP}  \Big)  :=  \widehat{\ell}(f_\matW(\vecx), y) \\
\le & \indi\Big( M(f_\matW(\vecx) , y) - \frac{\epsilon}{2} \max_{k\in[K] , z=\pm1} \max_{\matP\succeq 0, \diag(\matP) \le 1} \innerps{zQ(\vecw_{2,k}, \matW_1)}{\matP}  \le \gamma \Big),
\end{align*}
where the first inequality is due to the property of ramp loss, the second inequality is by the definition of the margin, the third inequality is due to Theorem~\ref{thm:sdp_surrogate}, the fourth inequality is due to~\eqref{eq:max_k_z}, the fifth inequality is by the definition of the margin and the last inequality is due to the property of ramp loss.

\subsection{Proof of Theorem~\ref{thm:sdp_gen}}\label{prf:sdp_gen}
We study the Rademacher complexity of the function class 
\[
\widehat{\ell}_\F := \{(\vecx, y) \mapsto \widehat{\ell}(f_\matW(\vecx), y) : f_\matW \in \F\}.
\]
Define $M_\F := \{(\vecx, y) \mapsto M(f_\matW(\vecx), y) : f_\matW \in \F\}$. Then we have
\begin{equation}\label{eq:split}
\frakR_\setS(\widehat{\ell}_\F) \le \frac{1}{\gamma} \Big( \frakR_\setS(M_\F) + \frac{\epsilon}{2n} \EE_{\vecsigma} \Big[ \sup_{f_\matW\in\F}\sum_{i=1}^n\sigma_i \max_{k\in[K] , z=\pm1} \max_{\matP\succeq 0, \diag(\matP) \le 1} \innerps{zQ(\vecw_{2,k}, \matW_1)}{\matP} \Big] \Big),
\end{equation}
where we use the Ledoux-Talagrand contraction inequality and the convexity of the supreme operation. For the first term, since we have $\| \matW_1 \|_{1} \le b_1$, we have $\| \matW_1^\top \|_{2,1} \le b_1$. Then, we can apply the Rademacher complexity bound in~\cite{bartlett2017spectrally} and obtain
\begin{equation}\label{eq:split_1}
\frakR_\setS(M_\F) \le \frac{4}{n^{3/2}} + \frac{60\log(n)\log(2d_{\max})}{n}s_1s_2 \left( (\frac{b_1}{s_1})^{2/3} + (\frac{b_2}{s_2})^{2/3} \right)^{3/2} \fbnorms{\matX}.
\end{equation}
Now consider the second term in~\eqref{eq:split}. According to~\cite{raghunathan2018certified}, we always have 
\begin{equation}\label{eq:max_positive}
\max_{\matP\succeq 0, \diag(\matP) \le 1} \innerps{zQ(\vecw_{2,k}, \matW_1)}{\matP} \ge 0.
\end{equation}
In addition, we know that when $\matP\succeq 0$ and $\diag(\matP) \le 1$, we have 
\begin{equation}\label{eq:norm_p}
\|\matP\|_\infty \le 1.
\end{equation}
Moreover, we have
\begin{equation}\label{eq:norm_w2}
\| \matW_2 \|_\infty \le \|\matW_2^\top\|_{2,1} \le b_2.
\end{equation}
Then, we obtain
\begin{align}
& \frac{\epsilon}{2n} \EE_{\vecsigma} \Big[ \sup_{f_\matW\in\F}\sum_{i=1}^n\sigma_i \max_{k\in[K] , z=\pm1} \max_{\matP\succeq 0, \diag(\matP) \le 1} \innerps{zQ(\vecw_{2,k}, \matW_1)}{\matP} \Big] \nonumber \\
\le & \frac{\epsilon}{2n} \Big( \sup_{f_\matW\in\F} \max_{k\in[K] , z=\pm1} \max_{\matP\succeq 0, \diag(\matP) \le 1} \innerps{zQ(\vecw_{2,k}, \matW_1)}{\matP}  \Big) \EE_{\vecsigma}\Big[ | \sum_{i=1}^n\sigma_i | \Big]  \nonumber \\
\le & \frac{\epsilon}{2\sqrt{n}}  \sup_{f_\matW\in\F} \max_{k\in[K] , z=\pm1} \max_{\matP\succeq 0, \diag(\matP) \le 1} \innerps{zQ(\vecw_{2,k}, \matW_1)}{\matP}  \nonumber  \\
\le &  \frac{\epsilon}{2\sqrt{n}} \sup_{f_\matW\in\F} \max_{k\in[K] , z=\pm1} \max_{\matP\succeq 0, \diag(\matP) \le 1} \|zQ(\vecw_{2,k}, \matW_1) \|_1 \|\matP\|_\infty  \nonumber \\
\le & \frac{2\epsilon}{\sqrt{n}} \sup_{f_\matW\in\F} \max_{k\in[K] } \onenms{\diag(\vecw_{2,k})^\top \matW_1}  \nonumber  \\
\le & \frac{2\epsilon}{\sqrt{n}} \sup_{f_\matW\in\F} \onenms{\matW_1} \| \matW_2 \|_\infty  \nonumber \\
\le & \frac{2\epsilon b_1b_2}{\sqrt{n}},  \label{eq:split_2}
\end{align}
where the first inequality is due to~\eqref{eq:max_positive}, the second inequality is due to Khintchine's inequality, the third inequality is due to H\"{o}lder's inequality, and the fourth inequality is due to the definition of $Q(\cdot, \cdot)$ and~\eqref{eq:norm_p}, the fifth inequality is a direct upper bound, and the last inequality is due to~\eqref{eq:norm_w2}.

Now we can combine~\eqref{eq:split_1} and~\eqref{eq:split_2} and get an upper bound for $\frakR_\setS(\widehat{\ell}_\F)$ in~\eqref{eq:split}. Then, Theorem~\ref{thm:sdp_gen} is a direct consequence of Theorem~\ref{thm:rad_vanilla} and Lemma~\ref{lem:surrogate}.
\end{document}